\pdfoutput=1

\documentclass[11pt]{article}

\usepackage{acl}
\usepackage{times}
\usepackage{latexsym}

\usepackage[T1]{fontenc}

\usepackage[utf8]{inputenc}

\usepackage{multirow}

\usepackage{microtype}
\usepackage{algorithm}
\usepackage{algpseudocode}
\usepackage{xcolor} 
\usepackage{inconsolata}
\usepackage{amsmath}
\usepackage{cleveref}
\usepackage{graphicx}

\usepackage{subcaption}
\usepackage{tabularx}
\usepackage{tikz}
\usepackage{amsfonts}
\graphicspath{{./images/}}
\newcolumntype{b}{X}
\newcolumntype{s}{>{\hsize=.5\hsize}X}

\usepackage[font=small,skip=0pt]{caption}

\hypersetup{
    colorlinks=true, 
    linkcolor=blue, 
    citecolor=purple, 
    urlcolor=blue 
}

\title{Large Language Models Relearn Removed Concepts}

\author{$\,\,\,\,\,\,\,\,\,\,\,\,\,\,\,\,\,\,\,\,\,\,\,\,$Michelle Lo$^{\heartsuit \spadesuit}$\Thanks{$\,\,$Equal contribution\\ Correspondence: \text{fazl@robots.ox.ac.uk}} $\qquad$ \\ 
\And
Shay B. Cohen$^\ddag$ \\ 
\And
Fazl Barez$^{\heartsuit \clubsuit *}$ \\
\AND
\textnormal{$^\heartsuit$ Apart Research}\\
\textnormal{$^\ddag$ School of Informatics, University of Edinburgh}\\
\textnormal{$^\spadesuit$ Department of Computing, Imperial College London}\\
\textnormal{$^\clubsuit$ Department of Engineering Sciences, University of Oxford}
}

\begin{document}
\maketitle
\begin{abstract}
Advances in model editing through neuron pruning hold promise for removing undesirable concepts from large language models. However, it remains unclear whether models have the capacity to reacquire pruned concepts after editing. To investigate this, we evaluate concept relearning in models by tracking concept saliency and similarity in pruned neurons during retraining. Our findings reveal that models can quickly regain performance post-pruning by relocating advanced concepts to earlier layers and reallocating pruned concepts to primed neurons with similar semantics. This demonstrates that models exhibit polysemantic capacities and can blend old and new concepts in individual neurons. While neuron pruning provides interpretability into model concepts, our results highlight the challenges of permanent concept removal for improved model \textit{safety}. Monitoring concept reemergence and developing techniques to mitigate relearning of unsafe concepts will be important directions for more robust model editing. Overall, our work strongly demonstrates the resilience and fluidity of concept representations in LLMs post concept removal.
\end{abstract}

\section{Introduction}
\label{sec:intro}

Large language models possess neurons that encode semantic concepts across different languages \cite{qian2016}, architectures \cite{wu2020}, and modalities \cite{muandreas2020}. 

The primary objective when pruning such models is to eliminate redundant neurons, while preserving the most crucial ones \cite{lecun1989}. The underlying assumption is that removing important ``concept neurons'' will disrupt the model's structured internal representation of key concepts, causing performance degradation. However, it is in fact possible for models to regain high performance after pruning random or important neurons \cite{Liu2022}. Models demonstrate a remarkable capacity to adapt and regain conceptual representations. This phenomenon, which we call \textbf{“neuroplasticity”}, suggests a degree of adaptability in such models.

Neuroplasticity has significant implications for model editing. Model editing promises to remove undesirable concepts, but neuroplasticity implies that those concepts may in fact reappear if retraining takes place. Understanding how concepts are represented, redistributed, and recaptured is thus crucial for the development of \textit{safer, fairer} and \textit{aligned} models. Furthermore, understanding how removed concepts are recovered in language models can significantly enhance their robustness. While the inherent plasticity of neural networks has been acknowledged \cite{mittal2019}, there is a limited understanding of where relearned concepts are redistributed, or what influences relearning in specific neurons. Such insights could benefit models' ability to recover from partial damage and information loss.

This paper investigates how large language models fine-tuned for named entity recognition relearn and redistribute concepts after the removal of important features, to explain how a model regains performance after experiencing damage to crucial neurons in its internal representation structure. We instigate neuroplasticity by pruning important concept neurons and retraining a model so that it regains its original performance. We then analyze how the distribution of concepts changes, and examine the relationship between concepts previously associated with a neuron and the concepts it relearns. Our findings are four-fold:

\begin{itemize}
    \item Neuroplasticity happens quickly, allowing the model to regain performance within a few epochs of retraining.
    \item Pruned concepts, initially present in later layers, are remapped to neurons in earlier layers.
    \item Neurons which recover the pruned concept may have been primed for relearning, as they originally captured concepts similar to that which was pruned.
    \item Neurons exhibit polysemantic properties as they relearn a blend of new and old concepts.
\end{itemize}

These findings contribute to a deeper understanding of how language models learn, adapt, and retain core conceptual representations. They underscore the potential of earlier layers to recapture fundamental representations, which has implications for model editing. Furthermore, our exploration of how neuroplasticity increases polysemanticity, where a neuron can represent multiple concepts and meanings may inform strategies to enhance the transfer of learned representations, and facilitate interpretability analysis regarding polysemanticity.

\section{Related work}
\label{sec:related-works}

In terms of analysing the distribution of concept representations, it is known that more complex concepts tend to be represented by neurons in higher layers of language models. \citet{durrani-etal-2020-analyzing} found that neurons which capture word morphology were primarily located in the lower and middle layers, and those learning about advanced concepts such as syntax were found at the higher layers. However, dynamic changes in concept distribution over training are less well studied. Prior works artificially redistributed concepts in a large language model by modifying the activations of specific neurons which capture desired properties, to control model outputs \cite{bau-et-al-2018}. However, these works focused on deliberately editing representations in a trained model. There is little work in which models redistribute concepts unsupervised, leaving a gap in understanding about how concept redistribution naturally occurs after pruning.

\begin{figure}[ht!]
\includegraphics[width=\linewidth]{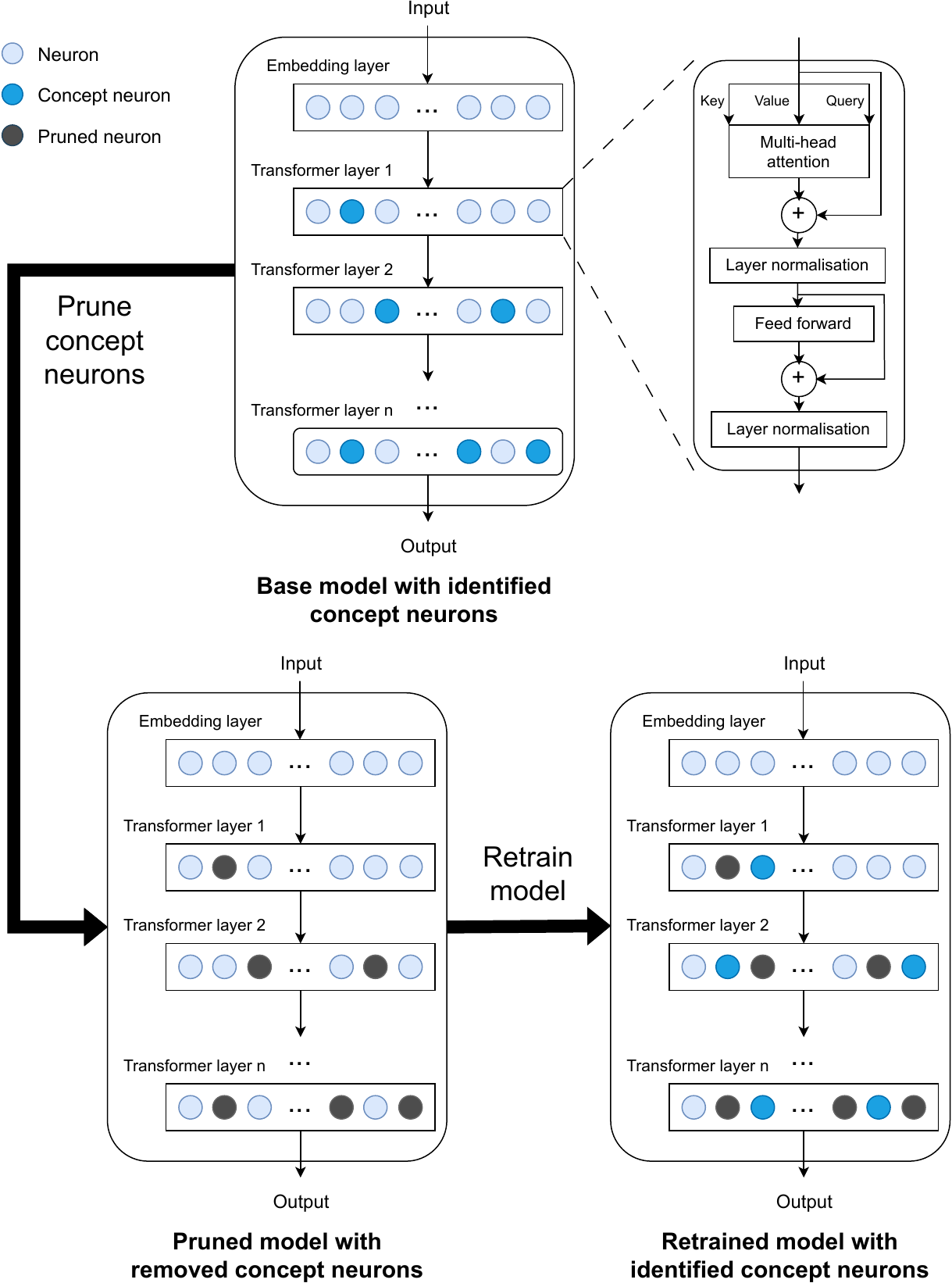}
\caption{Process of investigating neuroplasticity in a large language model. We identify concept neurons (dark blue) in the base model, and prune them (white). We then retrain the model until it regains its original performance and identify new concept neurons.}
\label{fig:overall-process}
\end{figure}

In terms of performance recovery after pruning, \citet{Liu2022} demonstrated that randomly pruned sparse networks can be trained from scratch to match the accuracy of equivalent dense networks, particularly in larger networks. Similarly, \citet{mittal2019} suggested that the inherent plasticity of deep convolutional neural networks allows them to recover from the loss of randomly pruned filters. Their work supports the idea that large neural networks demonstrate plasticity, and can regain performance if important neurons might be pruned. However, their approach uses indiscriminate random pruning of neurons in image models. In contrast, we explicitly prune the most important neurons, which allows us to analyze causation between removing crucial features and performance. In order to clearly analyze changes in the model before and after pruning, our approach also trains, prunes and retrains the model in separate phases - as opposed to other approaches, which iteratively prune and adjust weights \cite{castellano-fanelli}.

Our paper is most similar to \citet{dalvi2020}, in which neurons that are redundant with regards to a specific task are pruned from LLMs, and model performance is re-examined. However, \citet{dalvi2020} focus on identifying redundant neurons and do not damage key concept neurons. On the other hand, we focus on how the internal distribution of concept representations and model performance changes after pruning important neurons.

\section{Problem setting}
\label{sec:problem}

Consider a large language model $M$. Let $N$ be the set of all neurons in $M$. Consider a concept $C$ represented within $M$. 

A concept neuron is a neuron in $N$ which activates strongly on tokens associated with $C$. Let $S \subset N$ be the set of concept neurons in $M$ prior to ablation. Let the set of remapped neurons $R \subset N$ be the set of concept neurons associated with $C$ in $M$, after pruning neurons in $S$ and retraining. 


We analyze the redistribution of concepts in $M$ after neuroplasticity occurs, in terms of concept saliency and concept similarity.

\subsection{Concept saliency}
\label{sec:saliency}

Concept saliency is a measure for how strongly a neuron encodes a representation of a specific concept, compared to other neurons in the same model. A high saliency score (close to 1) means that the neuron captures the concept more strongly, relative to neurons in the rest of the model.

Concept saliency helps us address the question: \textit{where are recovered concepts distributed in $M$ when neuroplasticity occurs?} Concept saliency can be calculated for individual neurons, enabling fine-grained analysis, but the mean concept saliency for all neurons in a layer can also be used for layer-wise analysis. Analysis of concept saliency facilitates investigation into how a pruned concept is distributed before and after neuroplasticity.

To compute concept saliency scores, we obtain a global ordering of neurons, in which neurons are ranked according to the extent to which they activate on a given concept (see \S\ref{sec:probeless}). The ordered list of neurons is divided into groups of 100 consecutive neurons. The neurons in each group is assigned the same (normalised) saliency score, according to the group's position in the ordered list. For instance, a saliency score of 1.0 indicates that a neuron is one of the top 100 most salient neurons.

\subsection{Concept similarity}
\label{sec:similarity}

Concept similarity measures how similar the concept currently captured by a neuron is to the concept that it originally captured in the base model. Formally, if $C_r$ is the set of concepts represented by a remapped neuron $r \in R$ in $N$ after pruning and retraining, then $C_r'$ is concept associated with $r$ in $N$ \textit{before} pruning and retraining. For example, a neuron might initially be associated with names of animals ($C_r'$), but after pruning and retraining, it might be associated with names of places ($C_r$).

Similarity analysis helps us address the question: \textit{how do the concepts represented by specific neurons change during neuroplasticity?} It offers a quantitative measure to compare the concepts originally distributed in the base model, and the redistribution of concepts in the retrained model. This allows us to further interpret the model's internal representations and investigate potential relationships between neurons' previously acquired concepts and the recovered concept.

To compute concept similarity:

\begin{enumerate}
    \item Take $H$, the set of highest activating tokens for $R$ in $N$ before we induce neuroplasticity by pruning and retraining. 
    \item Take $H'$, the set of highest activating tokens for $R$ in $N$ after we induce neuroplasticity.
    \item Convert $H$ and $H'$ into sets of word embedding vectors $W$ and $W'$ respectively.
    \item Compute the cosine similarity between the average vectors of $W$ and $W'$.
\end{enumerate}

A similarity score close to 1 suggests that concepts are more semantically similar, and a score close to -1 indicates opposing concepts. If the similarity score is above a predetermined threshold (defined as 0.5 in this paper), we infer that there is a meaningful relationship between the newly learned concept $C_r$ and the concept that was originally captured $C_r'$ by the same neuron.

\section{Method}
\label{sec:method}
To explore the notion of neuroplasticity within a pretrained model $M$, we first fine-tune the model for a task that calls for the understanding of a specific concept $C$. Subsequently, we employ a probeless neuron search method to pinpoint the set of concept neurons $S$ within the model that correlate with $C$ (\S\ref{sec:probeless}). In order to remove the concept from the model's internal representations, we prune the neurons in $S$ (\S\ref{sec:pruning}). Following the pruning process, the model is retrained until it attains its original performance level. The recovery of performance signifies that neuroplasticity has taken effect, and that the eradicated concept has been relearned and integrated throughout the model. We can use the same probeless search method to determine the set of remapped neurons $R$ in the retrained model that have acquired the pruned concept. During the retraining period, we can examine how concept saliency and similarity evolve across neurons in both $S$ and $R$ at two-epoch intervals. The overall process of our methodology is shown in Figure \ref{fig:overall-process}. For detailed steps, refer to the pseudo-code algorithm \ref{alg:revised_algorithm} in Appendix \ref{app:algo}.


\subsection{Identifying top concept neurons}
\label{sec:probeless}

We use a probeless corpus-based neuron search method \cite{antverg2021} to build a list of top neurons with regards to a concept label. This list is used to compute concept saliency (\S\ref{sec:saliency}). We used the probeless search method implementation provided by the NeuroX toolkit \cite{dalvi2019neurox}.

Given labeled training data in which tokens are annotated with associated concepts, we use the NeuroX toolkit to extract neuron activations for every input token. Features include all unit activations on every token, for every neuron in the model. The probeless method ranks neurons according to the difference in their values across labeled tokens.

Formally, for every concept label $z \in Z$ where $Z$ is the set of all possible concepts label, we calculate $q(z)$, the mean vector of all representations of words that possess the concept label $z$. Next, we calculate the element-wise difference between the mean vectors for a specific concept $z$ ($d$ is the number of neuron activations):

\begin{equation}
\begin{aligned}
 r(z) = \sum_{z' \in Z, z \neq z'} |q(z) - q(z')|, \,\,r(z) \in \mathbb{R}^d\text{.}
\end{aligned}
\end{equation}

Following this, we obtain a ranking by arg-sorting the elements of $r(z)$. Then the first neuron in the ranking corresponds to the highest value in $r(z)$. A neuron's ranking is determined by the absolute variation between the mean activation of that neuron for a specific concept and its mean activation over all other concepts. A highly ranked neuron triggers more robustly on words related to a single concept, as compared to its average activation over all words. By following this calculation, we identify the neurons that behave the most differently from other neurons for concepts different than $z$.

Since the probeless method ranks neurons based on average neuron activations, this approach is capable of handling noisy activations in non-sparse representations. In instances where many units are activated for a specific concept, those with significantly ``large'' contributions will be identified as highly salient. Furthermore, the probeless method was chosen because it is based purely on representations, making it free of probing limitations that might affect ranking quality. The probeless method is also significantly faster than alternative methods such as probing classifiers, as we are only limited by averaging and sorting speeds.

\subsection{Pruning}
\label{sec:pruning}

To prune the concept neurons from the model, the weights of the selected neurons are set to zero. Formally, let $W$ be the model's weight matrix, and let $\Phi$ be a binary mask matrix indicating the selected neurons for pruning (with the same dimensions as $W$). Then, the pruned weight matrix $(W')$ is computed as: $W'= W \odot \Phi$ where $\odot$ denotes element-wise multiplication.

\section{Experimental setup}
\label{sec:experimental-setup}

We focus on pruning the specific concept of location names from a DistilBERT model, a DistilGPT2 model \cite{sanh2020distilbert} and a GPT2 model \cite{radford2019language} fine-tuned for named entity recognition. Analysis of the models across different runs are in Appendix \ref{sec:appendix}, and analysis of the models after pruning other concepts are in Appendix \ref{app:different-concept}.

\subsection{Model architecture}

\textbf{Base models: } We use the transformers library and the CoNLL-2003 dataset \cite{conll-2003} to fine-tune open-sourced, pretrained DistilBERT, DistilGPT2 and GPT-2 models for named entity recognition (NER). DistilBERT is a lighter version of the encoder-only BERT model with fewer layers and parameters (66 million parameters), maintaining competitive performance while requiring less compute resources \cite{sanh2020distilbert}. GPT-2 is a large causal, decoder-only transformer with 1.5 billion parameters \cite{radford2019language}. DistilGPT2 is a faster, lighter version of GPT-2 with 82 million parameters \cite{sanh2020distilbert}. Since DistilBERT, DistilGPT2 and GPT-2 represent LLMs with different architectures and training objectives, analysis on these models are more \textit{generalizable} to other LLMs.

\textbf{Training:}
\label{model_train}
Models are fine-tuned for three epochs using AdamW optimizer with a linear scheduler and warmup \cite{loshchilov2019decoupled}. The hyperparameters used are: learning rate of 2e-5, weight decay of 0.01, batch size of 8.
\textbf{Evaluations:}
Entity recognition performance is evaluated using precision, recall, and F1 computed with the seqeval library \cite{seqeval}. We perform cross-validation, where we examine the highest activating tokens for each fold to validate the robustness of the recognized entities.


\subsection{Identifying concept neurons}

To detect top concept neurons, we use concept-annotated tokens from the dataset BERT ConceptNet \cite{dalvi2020discovering}. BERT ConceptNet is a dataset of latent concepts learned within the representations of BERT. The dataset consists of sets of sentences, for which each token in a sentence is annotated with its associated concept. Tokens are classified hierarchically based on their semantic properties, e.g. tokens associated with location have the label SEM:named\_entity:location. As we focus on identifying salient neurons for a particular concept, it is feasible to view this as a binary classification problem over 1) tokens annotated with the chosen concept, and 2) tokens that are not annotated as related to the selected concept.

\subsection{Pruning and retraining}

We use the probeless method (\S\ref{sec:probeless}) to extract a list of all neurons ordered by concept saliency. We verify our measure of concept saliency corresponds to the extent to which a concept is captured in Appendix \ref{app:verify-saliency}. We prune the the most salient half of the model's neurons (i.e. neurons in the top half of the ordered list of most salient neurons).


After pruning, we retrain the model using details in \S\ref{model_train}. After every three epochs, we save the model's state so that we can analyze the pattern of concept redistribution over the retraining process incrementally. We compare the performance of the retrained model against the base model in terms of precision, F1 and recall.

\subsection{Extracting highest activating tokens}

To extract the highest activating tokens (HATs) for a neuron, we use the NeuroX library \cite{dalvi2019neurox} and compare the activations of the neuron across all tokens in a given dataset of sentences. The top five tokens that account for the largest variance in activations form the set of HATs for that neuron. Each HAT has a corresponding activation score, which is a normalised score depicting its contribution to the overall variance. We use Word2Vec \cite{mikolov2013word2vec} to convert HATs to word embeddings, to compute concept similarity (\S\ref{sec:similarity}).




\section{Results}

In our results, performance metrics indicate that neuroplasticity occurs early in the retraining process (\S\ref{section:performance}). Specifically, neuroplasticity occurs when pruned concepts are relocated to earlier layers in a model (\S\ref{section:redistribution}). This may be because neurons which reacquire the pruned concept are primed for relearning, having captured similar concepts initially (\S\ref{section:similar-concepts}). Concept similarity scores show that the pruned concept is relearned in addition to original concepts, in a polysemantic manner (\S\ref{section:polysemantic}).

\subsection{Random pruning baseline}
To verify that the neuroplasticity effects are specifically linked to pruning critical concept neurons, we establish a random pruning baseline where the same number of neurons are removed randomly instead of targeting concept neurons (Table \ref{table:concept_random-gpt2}). After random pruning, the model experiences a more drastic performative drop on NER compared to concept pruning (F1 decreases from 0.443 to 0.219). Recovery during retraining is also slower, reaching only 0.330 F1 after two epochs. This aligns with expectations, as random neurons likely do not capture coherent concepts critical for NER.

However, with continued retraining, model performance matches and slightly exceeds the original base model, implying that representation remapping does occur but in a less directed manner. Analyzing the random pruning baseline helps confirm that the rapid, adaptive plasticity behaviors are unique outcomes of pruning semantic concept neurons rather than pruning arbitrary neurons. Concept redistribution trends also differ between random pruning versus concept pruning. As seen in Table \ref{tab:distribution_random-bert}, after random pruning, while later layers gain more salient neurons, earlier and middle layers do not gain many more salient neurons compared to concept pruning. Hence, the key difference is that the earlier/middle layers do not capture the pruned concept as strongly with random pruning compared to concept pruning.

\subsection{Rapid performance recovery after retraining}
\label{section:performance}

\begin{table}[htb!]
\resizebox{\linewidth}{!}{%
\begin{tabular}{p{0.23\linewidth}rrrrrr}
\hline
\multirow{2}{*}{\textbf{Stage}} &
  \multicolumn{2}{l}{\textbf{Precision}} &
  \multicolumn{2}{l}{\textbf{Recall}} &
  \multicolumn{2}{l}{\textbf{F1}} \\
 &
  \multicolumn{1}{l}{\textit{Mean}} &
  \multicolumn{1}{l}{\textit{SD}} &
  \multicolumn{1}{l}{\textit{Mean}} &
  \multicolumn{1}{l}{\textit{SD}} &
  \multicolumn{1}{l}{\textit{Mean}} &
  \multicolumn{1}{l}{\textit{SD}} \\
\hline
Base model           & 0.398& 0.007& 0.499& 0.008& 0.443& 0.001\\
\textbf{Pruned}      & \textbf{0.210}& 0.007& \textbf{0.229}& 0.023& \textbf{0.219}& 0.014\\
Retrained (2 epochs) & 0.297& 0.010& 0.370& 0.035& 0.330& 0.020\\
Retrained (4 epochs) & 0.363& 0.008& 0.452& 0.017& 0.402& 0.002\\
Retrained (6 epochs) & 0.393& 0.013& 0.490& 0.006& 0.436& 0.011\\
Retrained (8 epochs) & 0.403& 0.009& 0.517& 0.004& 0.453& 0.007\\
\hline
\end{tabular}%
}
\caption{Means and standard deviations (SD) to 3 d.p. of precision, recall and F1 score on overall NER, after \textit{randomly} pruning neurons from DistilGPT2 over 3 runs.}
\label{table:concept_random-gpt2}
\end{table}

\begin{table}[htb]
\resizebox{\linewidth}{!}{%
\begin{tabular}{p{0.23\linewidth}rrrrrr}
\hline
\multirow{2}{*}{\textbf{Stage}} &
  \multicolumn{2}{l}{\textbf{Precision}} &
  \multicolumn{2}{l}{\textbf{Recall}} &
  \multicolumn{2}{l}{\textbf{F1}} \\
 &
  \multicolumn{1}{l}{\textit{Mean}} &
  \multicolumn{1}{l}{\textit{SD}} &
  \multicolumn{1}{l}{\textit{Mean}} &
  \multicolumn{1}{l}{\textit{SD}} &
  \multicolumn{1}{l}{\textit{Mean}} &
  \multicolumn{1}{l}{\textit{SD}} \\
\hline
Base model           & 0.932 & 0.002 & 0.940 & 0.001 & 0.937 & 0.001 \\
\textbf{Pruned}      & \textbf{0.019} & 0.003 & \textbf{0.147} & 0.015 & \textbf{0.034} & 0.005 \\
Retrained (2 epochs) & 0.909 & 0.013 & 0.922 & 0.011 & 0.915 & 0.011 \\
Retrained (4 epochs) & 0.918 & 0.008 & 0.930 & 0.008 & 0.924 & 0.008 \\
Retrained (6 epochs) & 0.919 & 0.007 & 0.932 & 0.003 & 0.926 & 0.006 \\
Retrained (8 epochs) & 0.922 & 0.008 & 0.934 & 0.006 & 0.928 & 0.007 \\
\hline
\end{tabular}%
}
\caption{Means and standard deviations (SDs) to 3 d.p. of precision, recall and F1 score on overall NER throughout remapping, after pruning concept neurons from DistilBERT over 3 runs. The dramatic drop in performance after pruning concept neurons confirms we removed important features.}
\label{table:concept}
\end{table}

\begin{table}[htb!]
\resizebox{\linewidth}{!}{%
\begin{tabular}{p{0.23\linewidth}rrrrrr}
\hline
\multirow{2}{*}{\textbf{Stage}} &
  \multicolumn{2}{l}{\textbf{Precision}} &
  \multicolumn{2}{l}{\textbf{Recall}} &
  \multicolumn{2}{l}{\textbf{F1}} \\
 &
  \multicolumn{1}{l}{\textit{Mean}} &
  \multicolumn{1}{l}{\textit{SD}} &
  \multicolumn{1}{l}{\textit{Mean}} &
  \multicolumn{1}{l}{\textit{SD}} &
  \multicolumn{1}{l}{\textit{Mean}} &
  \multicolumn{1}{l}{\textit{SD}} \\
\hline
Base model           & 0.394& 0.005& 0.483& 0.002& 0.434& 0.003\\
\textbf{Pruned}      & \textbf{0.363}& 0.004& \textbf{0.410}& 0.012& \textbf{0.385}& 0.008\\
Retrained (2 epochs) & 0.415& 0.003& 0.500& 0.009& 0.454& 0.005\\
Retrained (4 epochs) & 0.428& 0.009& 0.527& 0.010& 0.472& 0.010\\
Retrained (6 epochs) & 0.443& 0.002& 0.569& 0.012& 0.495& 0.000\\
Retrained (8 epochs) & 0.458& 0.004& 0.568& 0.005& 0.507& 0.005\\
\hline
\end{tabular}%
}
\caption{Means and standard deviations (SD) to 3 d.p. of precision, recall and F1 score on overall NER over remapping, after pruning concept neurons from DistilGPT2 over 3 runs. The drop in performance after pruning concept neurons confirms we have removed important features.}
\label{table:concept-gpt2}
\end{table}

Upon removing the top concept neurons, there is a substantial decline in a model's performance on sequence evaluation tasks. For DistilBERT, there is a considerable decrease in precision (by 98\%), recall (by 84\%), and F1 score (by 96\%) as shown in Table \ref{table:concept}. Notably, the performance drop for named entity recognition (NER) regarding location names is even more pronounced - with a 95.9\% decrease in precision, 93\% decrease in recall, and 94.9\% decrease in F1 score.
This is expected. Such drastic declines verify that we have removed the most important neurons used by the model for NER. For DistilGPT2, there is also a considerable decrease across all scores (Table \ref{table:concept-gpt2}). Absolute performance scores for DistilGPT2 are worse than DistilBERT - this is also expected, as DistilGPT2 is a decoder-only model that is less suited for sequence evaluation tasks such as named entity recognition.

When highly significant neurons are removed, the model's capacity to grasp critical concepts recovers after retraining. The pruned model exhibits a marked enhancement in performance within only two epochs. For example, the difference between the DistilBERT model's mean F1 score before pruning and after retraining for two epochs is merely 0.022. Over eight epochs of retraining, the model progressively improves in precision, recall, and F1 score, until performance matches or exceeds the baseline.

The sudden jump after two epochs of retraining and the subsequent plateau in performance indicates that neuroplasticity happens quickly in the early stages of retraining. In the next section, we demonstrate that this can be attributed to the redistribution of the pruned concepts to remaining neurons in earlier layers of the model. 


\subsection{High level concept redistribution}
\label{section:redistribution}

Concepts which were originally located in later layers are redistributed over both later and middle layers of the model during neuroplasticity. This is evident from 1) changes in the proportion of salient neurons per layer, and 2) changes in mean saliency per layer across stages of retraining.

First, consider the proportion of salient neurons per layer. Table \ref{tab:distribution-bert} and Table \ref{tab:distribution-gpt2} show the distribution of salient neurons across layers before and after inducing neuroplasticity, in a DistilBERT model and a DistilGPT2 model respectively. Specifically, these tables show the percentage of neurons in a given layer which have a concept saliency greater than 0.5, normalized over the number of non-pruned neurons in the given layer.

Table \ref{tab:distribution-bert}, Table \ref{tab:distribution-distilgpt2} and Table \ref{tab:distribution-gpt2} indicate that later layers initially have the highest concentration of representations for a pruned concept, but concept saliency becomes concentrated in either middle or later layers of the model after neuroplasticity. Before pruning concept neurons and retraining, the last layer of both models has the highest percentage of salient neurons. After ablating concept neurons and retraining the models, the pruned concept is redistributed such that layers located earlier in DistilBERT and DistilGPT2 model (layer 4 and layer 5 respectively) shows higher concentrations of salient neurons. For GPT-2, however, concept saliency redistributes more evenly across the model after pruning and retraining. This could be due to differences in how layers in larger models respond to inputs and tokens.

\begin{table}[t]
    \centering
    \resizebox{\linewidth}{!}{%
    \begin{tabular}{ccc}
    \hline
 & \multicolumn{2}{c}{\textbf{Percentage of salient neurons in layer}}\\
         \textbf{Layer} & \textbf{Base model}& \textbf{Pruned and retrained}\\
         \hline
         0&  0.111& 3.308\\
         1&  15.481& 5.923\\
         2&  17.778& 14.077\\
         3&  19.259& 15.615\\
         4&  19.333& 17.462\\
         5&  \textbf{21.000}& 15.308\\
         6&  7.037& \textbf{28.308}\\
         \hline
    \end{tabular}%
    }
    \caption{Normalized percentage of neurons which demonstrate concept saliency, before and after baseline DistilGPT2 model is \textit{randomly} pruned and retrained.}
    \label{tab:distribution_random-bert}
\end{table}

\begin{table}[t]
    \centering
    \resizebox{\linewidth}{!}{%
    \begin{tabular}{ccc}
    \hline
 & \multicolumn{2}{c}{\textbf{Percentage of salient neurons in layer}}\\
         \textbf{Layer} & \textbf{Base model}& \textbf{Pruned and retrained}\\
         \hline
         0&  12.28& 8.98\\
         1&  11.36& 13.47\\
         2&  10.35& 16.25\\
         3&  11.41& 16.25\\
         4&  13.51& \textbf{20.11}\\
         5&  19.01& 15.98\\
         6&  \textbf{22.08}& 8.98\\
         \hline
    \end{tabular}%
    }
    \caption{Normalized percentage of neurons which demonstrate concept saliency, before and after the DistilBERT model is pruned and retrained. Layer 6 is the most concept salient prior to ablation, but layer 4 is the most concept salient after ablation and retraining.}
    \label{tab:distribution-bert}
\end{table}

\begin{table}[h!]
    \centering
    \resizebox{\linewidth}{!}{%
    \begin{tabular}{clc}
    \hline
     &\multicolumn{2}{c}{\textbf{Percentage of salient neurons in layer}}\\
         \textbf{Layer} &  \textbf{Base model}&\textbf{Pruned and retrained}\\
         \hline
         0&  1.000&0.037\\
         1&  15.846&14.963\\
         2&  14.923&17.704\\
         3&  13.692&19.074\\
         4&  14.154&19.741\\
         5&  14.538&\textbf{20.704}\\
         6& \textbf{25.846}&7.778\\
         \hline
    \end{tabular}%
    }
    \caption{Normalized percentage of neurons which demonstrate concept saliency, before and after the DistilGPT2 model is pruned and retrained. Concept saliency increases in layers 4 and 5 after retraining, and decreases dramatically for layer 6.}
    \label{tab:distribution-distilgpt2}
\end{table}

\begin{table}
    \centering
    \resizebox{\linewidth}{!}{%
    \begin{tabular}{clc}
    \hline
     &\multicolumn{2}{c}{\textbf{Percentage of salient neurons in layer}}\\
         \textbf{Layer} &  \textbf{Base model}&\textbf{Pruned and retrained}\\
         \hline
         0&  0.000&0.217\\
         1&  5.587&6.000\\
         2&  6.696&8.043\\
         3&  7.587&9.739\\
         4&  8.304&8.826\\
         5&  8.935&8.391\\
         6& 9.152&8.696\\
         7& 9.696&8.304\\
         8& 10.326&7.478\\
         9& 10.652&7.522\\
         10& 10.891&7.870\\
         11& \textbf{11.391}&7.696\\
         12& 0.783&11.217\\
         \hline
    \end{tabular}%
    }
    \caption{Normalized percentage of neurons which demonstrate concept saliency, before and after a GPT2 model is pruned and retrained. Most salient neurons initially appear in later layers, but concept saliency redistributes approximately evenly.}
    \label{tab:distribution-gpt2}
\end{table}

\begin{figure}[htb!]
\includegraphics[width=\linewidth,trim={0cm 0cm 0cm 1.5cm},clip]{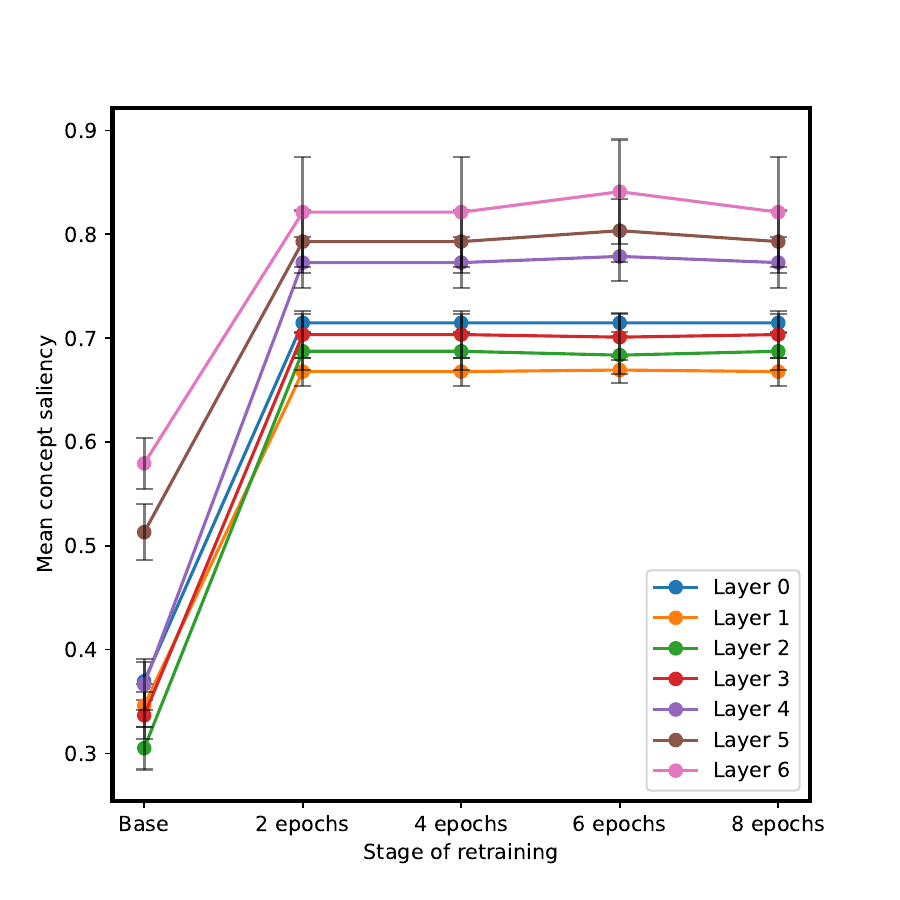}
\caption{Mean concept saliency for the concept of location names, for neurons across different layers of DistilBERT throughout the process of neuroplasticity. The most salient neurons are in layers 5 and 6, but earlier layers demonstrate higher saliency scores relative to other layers than before.}
\label{fig:saliency-layers-bert}
\end{figure}

\begin{figure}[htp]
\includegraphics[width=\linewidth]{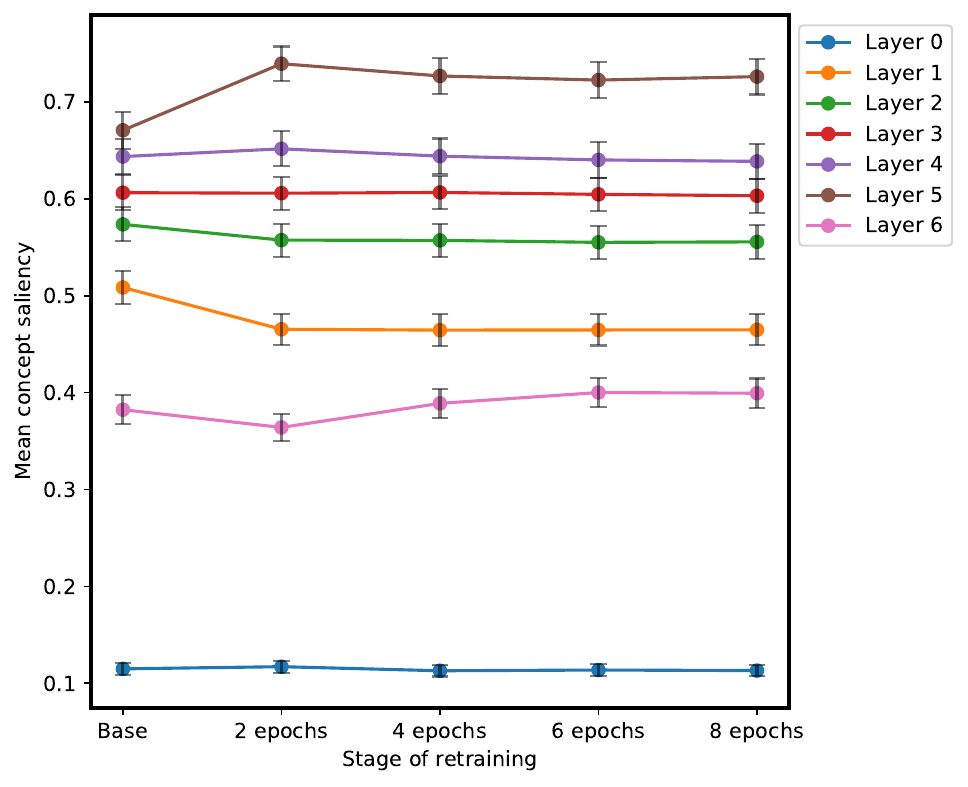}
\caption{Mean concept saliency for the concept of location names, for neurons across different layers of DistilGPT2 throughout the process of neuroplasticity. Saliency increases for middle and later layers (layers 5 and 6).}
\label{fig:saliency-layers-gpt2}
\end{figure}

\begin{figure}[htb!]
\includegraphics[width=\linewidth]{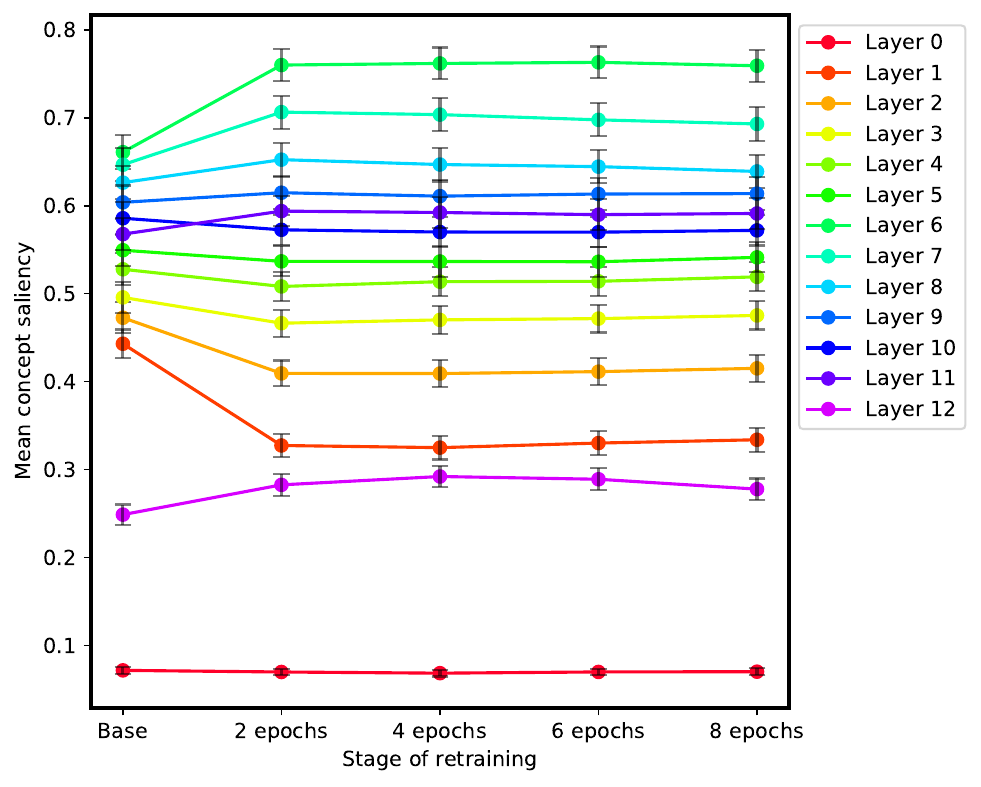}
\caption{Mean concept saliency for the concept of location names, for neurons across different layers of GPT2 throughout neuroplasticity. Results are similar to those for DistilGPT2.}
\label{fig:saliency-layers-fullgpt2}
\end{figure}

Next, consider the mean saliency scores per layer after ablation and throughout retraining. Figure \ref{fig:saliency-layers-bert} and Figure \ref{fig:saliency-layers-gpt2} show that mean concept saliency score increases after pruning and retraining for middle layers, more than for later layers. In DistilBERT, the increase in mean concept saliency for layer 4 is greater than the increase in mean concept saliency for layers 5 and 6, which originally had the most salient neurons (Figure \ref{fig:saliency-layers-bert}). In DistilGPT2, the increase in mean concept saliency for layer 5 is greater than the increase in mean concept saliency for layer 6. Other layers do not change significantly in saliency (Figure \ref{fig:saliency-layers-gpt2}). The earliest layers in DistilGPT2 even experience a decrease in mean concept saliency, suggesting that concept saliency becomes especially concentrated in later and middle layers. Results for DistilGPT2 are reflected in results for GPT-2 (Figure \ref{fig:saliency-layers-fullgpt2}).

\begin{figure}[htb!]
\includegraphics[width=\linewidth]{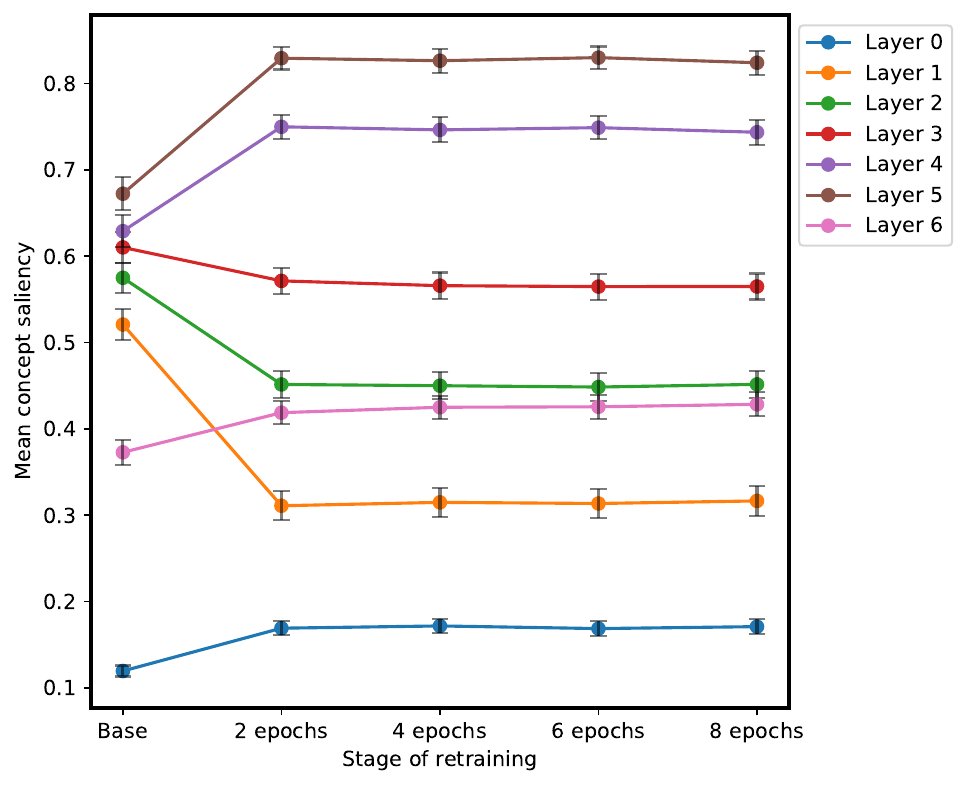}
\caption{Mean concept saliency for the concept of location names, for neurons across different layers of a baseline DistilGPT2 model throughout the process of neuroplasticity, after pruning random neurons.}
\label{fig:saliency-layers-baseline}
\end{figure}

For more concrete examples based on highest activation tokens of how concept saliency is redistributed to earlier layers, refer to Appendix \ref{app:redistribution}.



\subsection{Pruned concept relocation}
\label{section:similar-concepts}

One explanation for why pruned concepts are redistributed to middle layers is that these neurons are ``primed'' for relearning, since they originally captured similar concepts to the pruned concept.

\begin{figure}[htb]
\includegraphics[width=\linewidth,trim={0.5cm 0cm 0.5cm 1.4cm},clip]{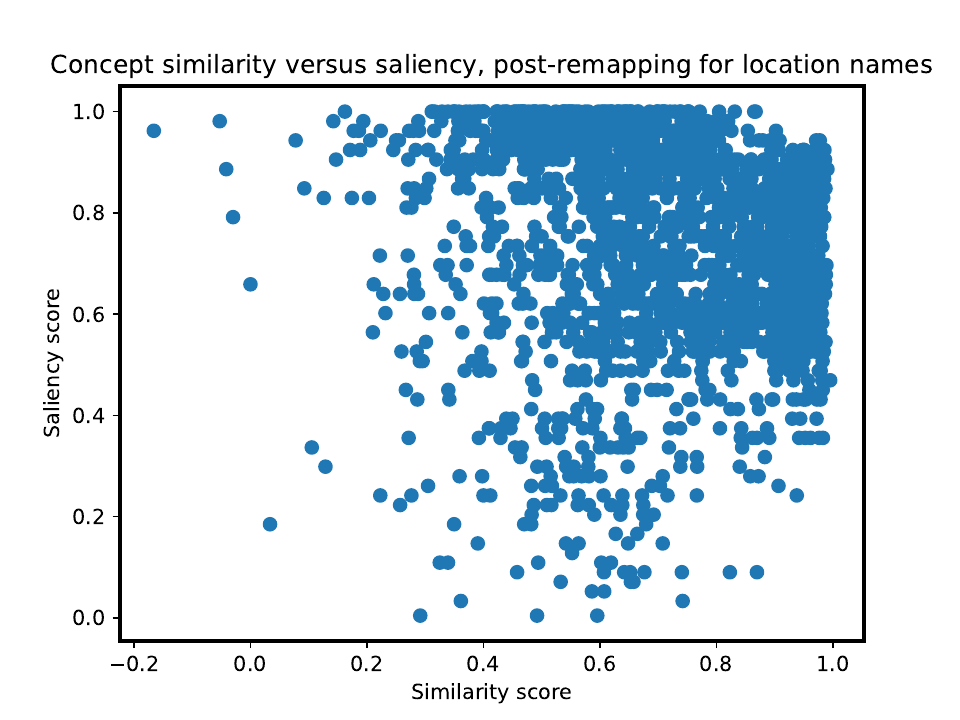}
\caption{In a DistilBERT model which was retrained for 8 epochs after the concept of location names was pruned, neurons which have strongly relearned the pruned concept (high saliency) tend to have previously represented a similar concept (high similarity) before pruning.
}
\label{fig:similarity-saliency-location}
\end{figure}

\begin{figure}[htb]
\includegraphics[width=\linewidth,trim={0.5cm 0cm 0.5cm 1.4cm},clip]{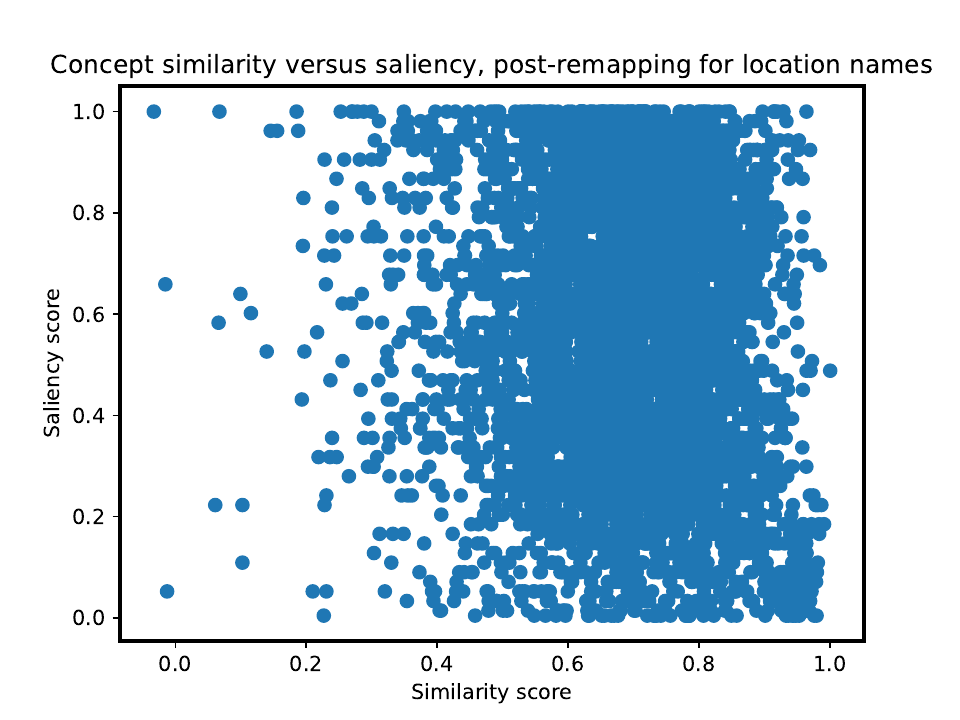}
\caption{In a DistilGPT2 model which was retrained for 8 epochs after the concept of location names was pruned, most neurons capture concepts after pruning and retraining that are similar to previously represented concepts - regardless of saliency.
}
\label{fig:similarity-saliency-gpt2}
\end{figure}

In DistilGPT2 models after pruning and retraining, most concept similarity scores are greater than 0.5 (Figure \ref{fig:similarity-saliency-gpt2}). This shows similar concepts are represented in neurons before and after pruning and retraining. In DistilBERT models, Figure \ref{fig:similarity-saliency-location} suggests that neurons which relearned the pruned concept (i.e. have high concept saliency scores) tend to have also previously captured similar concepts. The correlation between saliency and similarity is replicated when pruning different concepts in DistilBERT models (Appendix \ref{app:person-names}). In other words, the pruned concept is relearned by neurons which previously captured representations of similar concepts. See Appendix \ref{app:name-saliency-similarity} for further analysis.

\subsection{Polysemantic characteristics}
\label{section:polysemantic}


Pruned concepts are redistributed to neurons that previously captured similar concepts, such that neurons relearn both aspects of the concept that they initially captured as well as the pruned concept. In other words, neurons in large language models become polysemantic after retraining and capture several concepts at once \cite{xin2019}.

\begin{figure}[htb!]
\includegraphics[width=\linewidth]{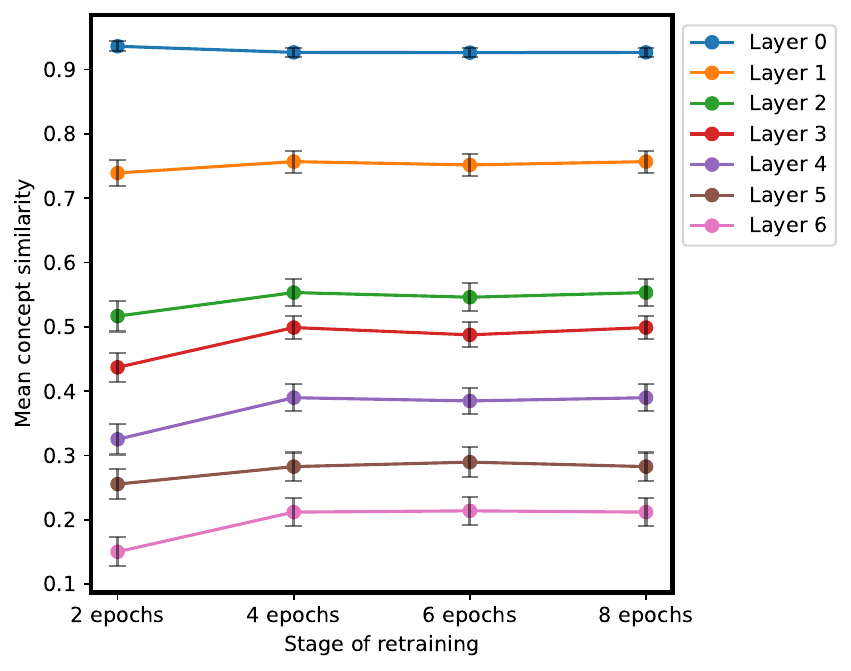}
\caption{Mean concept similarity for the concept of location names, for different layers of a DistilBERT model during neuroplasticity. Error bars represent 95\% confidence intervals. Concept similarity increases slightly in later layers (4, 5, 6) as retraining proceeds. Neurons in these layers are relearning old concepts during neuroplasticity.}
\label{fig:similarity-graph-bert}

\end{figure}
\begin{figure}[htb!]
\includegraphics[width=\linewidth]{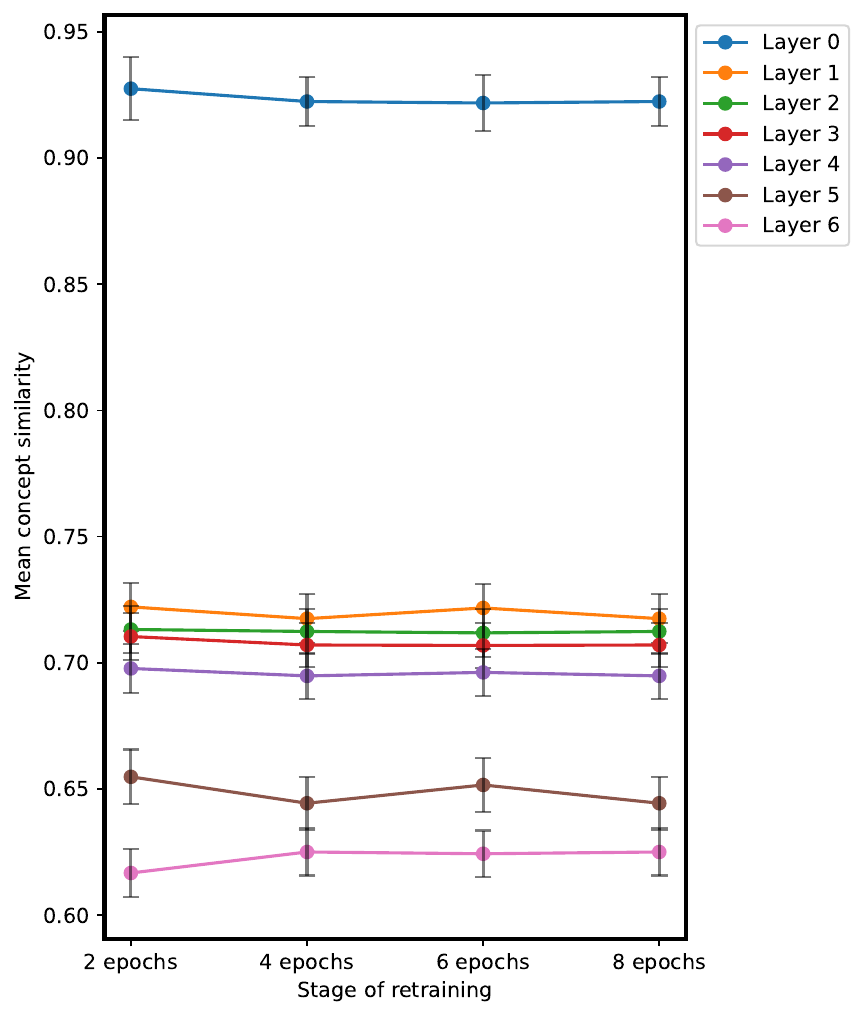}
\caption{Mean concept similarity for the concept of location names, for different layers of a DistilGPT2 model during neuroplasticity. Error bars represent 95\% confidence intervals. Concept similarity stays stable with minor fluctuations.}
\label{fig:similarity-graph-distilgpt2}
\end{figure}
\begin{figure}[htb!]
\includegraphics[width=\linewidth]{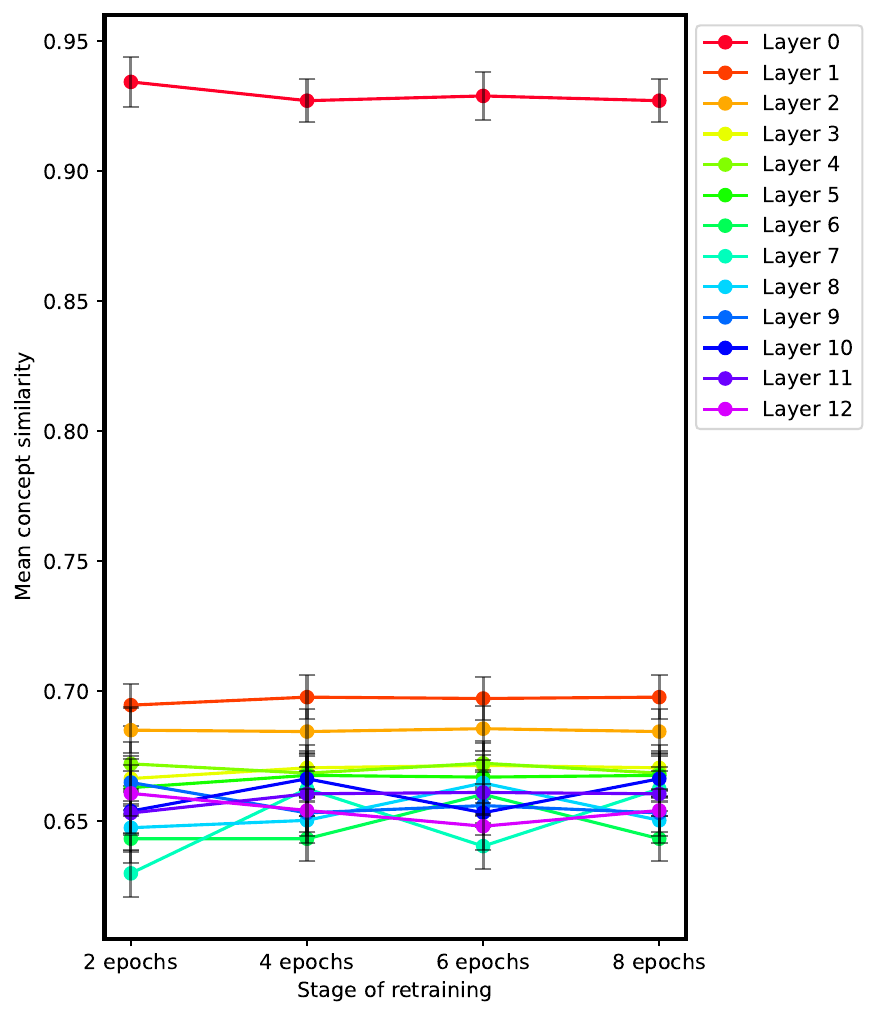}
\caption{Mean concept similarity for the concept of location names, for different layers of a GPT2 model during neuroplasticity. Results are similar to those for DistilGPT2, indicating generalization.}
\label{fig:similarity-graph-gpt2}
\end{figure}

Throughout retraining, neurons adapt to capture concepts that are increasingly similar to the concepts that they originally captured prior to ablation. This appears to be especially true for encoder-only models such as DistilBERT. Layers which adapt to relearn the pruned concept in DistilBERT (layer 4) experience a greater increase in mean concept similarity scores, compared to other layers in the model (Figure \ref{fig:similarity-graph-bert}). Earlier layers learn concepts that are much more similar to concepts that are captured in the base model, and remain stable in concept similarity scores throughout retraining. In contrast, the mean similarity for later layer neurons in DistilBERT fluctuates more and rises more as retraining progresses. In DistilGPT2, concept similarity stays stable with minor fluctuations over retraining (Figure \ref{fig:similarity-graph-distilgpt2}). Results for DistilGPT2 are reflected in results for GPT-2 (Figure \ref{fig:similarity-graph-gpt2}). This indicates that, in addition to learning to represent the pruned concept, later layers relearn to represent aspects of originally captured concepts.

\begin{figure}[htb!]
\includegraphics[width=\linewidth]{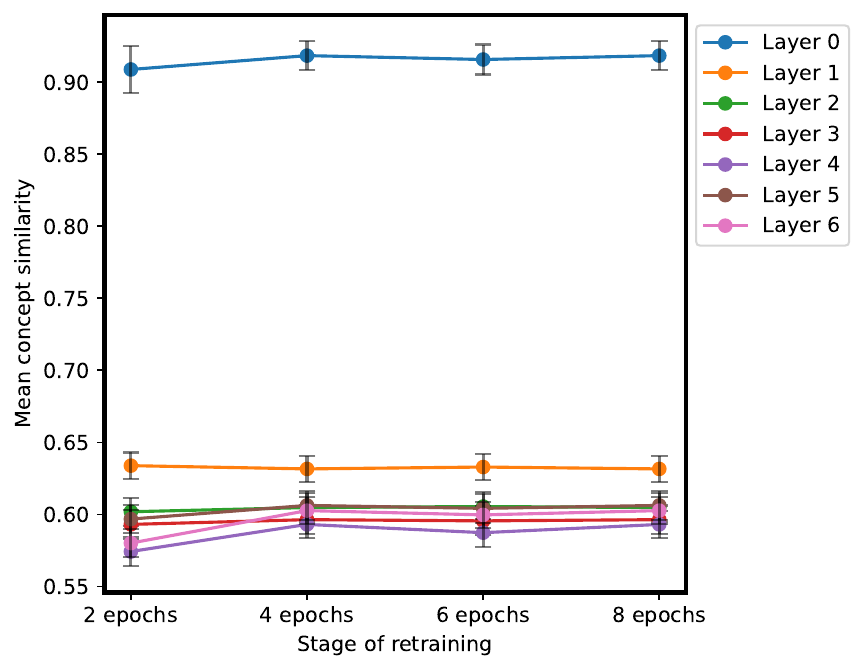}
\caption{Mean concept similarity for the concept of location names, for different layers of a baseline (randomly pruned) DistilGPT2 model during neuroplasticity.}
\label{fig:similarity-graph-baseline}
\end{figure}

\begin{table}[!htb]
\centering
\resizebox{\linewidth}{!}{%
\begin{tabular}{p{0.22\linewidth}p{0.8\linewidth}}
\hline
\textbf{Stage} & \textbf{HATs and activations}\\
\hline
Base model & ("1970", 1.0), (\textbf{"Bronze"}, 0.960), ("Administrative", 0.936), ("Thoroughbred", 0.913), (\textbf{"Egyptian"}, 0.899) \\
Retrained (2 epochs) & ("Palestinian", 1.0), ("Maltese", 0.998), ("Egyptian", 0.983), ("Venezuelan", 0.964), ("Damascus", 0.959) \\
Retrained (4 epochs) & ("Maltese", 1.0), ("Palestinian", 0.956), ("Nuremberg", 0.940), ("Egyptian", 0.934), ("Nordic", 0.927) \\
Retrained (6 epochs) & ("Maltese", 1.0), ("Nuremberg", 0.978), ("Malaysian", 0.955), ("Nordic", 0.948), ("Palestinian", 0.944) \\
Retrained (8 epochs) & ("Maltese", 1.0), ("Malaysian", 0.944), (\textbf{"Bronze"}, 0.934), (\textbf{"Egyptian"}, 0.929), ("Nordic", 0.929)\\
\hline
\end{tabular}%
}
\caption{Highest activating tokens for neuron 4148 in DistilBERT throughout stages of neuroplasticity, for the concept of location names. The neuron learns to activate on both new tokens related to location, and old tokens (in bold) related to its original concept captured in the base model.}
\label{tab:hats-polysem}
\end{table}

Analysis of the HATs for individual neurons provides concrete examples of how both new and old concepts are captured after neuroplasticity occurs. In one such example (Table \ref{tab:hats-polysem}), neuron 4148 in layer 5 of a DistilBERT model learns to activate on tokens (e.g. "Damascus") related to the pruned concept of location names after retraining, when it previously did not do so. Over 8 epochs of retraining, the neuron not only relearns new concepts related to location ("Maltese", "Malaysian") but also relearns concepts that it activated highly on in the base model ("Bronze", "Egyptian"). The polysemantic neuron after retraining captures aspects of both new and old concepts, although the newly-learned pruned concept is represented more strongly (as seen by higher activation scores). These results are replicated for different neurons in Appendix \ref{app:name-remapping-hats}. 

\section{Conclusion}

We explored neuroplasticity in large language models (DistilBERT, DistilGPT2 and GPT2) and implications on the distribution of concept representations. Our analysis showed that these models redistribute concepts after the removal of important features, allowing them to regain performance within a few epochs of retraining. The recovery in performance can be attributed to how pruned concepts, initially present in later layers, are remapped to remaining neurons in earlier layers which capture similar concepts. We further identified that neurons exhibit polysemantic properties as they relearn a blend of new and old concepts.

These insights contribute to a deeper understanding of how language models learn, adapt, and retain core conceptual representations. They also suggest interesting research directions in model editing and transfer learning. For instance, models may need to be edited throughout retraining to ensure that pruned concepts are not recaptured by earlier layers during fine-tuning. Studying the implications of neuroplasticity-induced polysemanticity can also aid the development of interpretable models and the enhanced transfer of learned representations.

\section*{Limitations}

In this paper, we only evaluate neuroplasticity in DistilBERT, DistilGPT2 and GPT-2 models. We expect that our overall findings regarding high level concept redistribution are generalizable, according to the similarity of results between DistilGPT2 and GPT-2, and because different architectures are know to have rather similar representations \cite{wu2020}. However, it is possible that language models with different sizes and architectures may recover performance and relearn concepts differently. We leave this exploration for future work.

More specifically, in \S\ref{section:similar-concepts}, we show that pruned concepts primarily relocate to neurons that previously captured similar concepts. However, we must emphasise that the precise relationship between the two factors is not clear. Consider figures \ref{fig:similarity-saliency-location} and \ref{fig:similarity-saliency-person}. In both cases, most data points are clustered in the region above the threshold of 0.5 for both saliency and similarity, and there is a large range in saliency and similarity within this region. The standard deviations for both concept saliency and concept similarity are relatively large (0.205 and 0.199, respectively). 
Hence, without further investigation, we can only state that pruned concepts \textit{tend} to relocate to neurons that previously captured similar concepts, but we must be cautious of drawing definite links. 

Finally, we compute concept saliency using a probeless method to extract a global ranking of neurons, ordered by the extent to which a neuron activates on a given concept. To verify the correctness of this method, we cross-checked our overall saliency scores with individual neurons - but our method is not scalable if one wishes to carefully analyze many neurons at once. We are also aware of the wider implications of this work and have added a broader impact statement in Appendix \ref{app:sub-app-brd}.

\section*{Acknowledgements}
We'd like to thank Mrinank Sharma and Clement Neo for the generous comments and feedback on the earlier draft and Esben Kran of Apart Research for organising the Interpretability hackathon. This project was supported by Apart Lab.

\bibliography{emnlp2023}

\appendix

\section{Experimental details}
\subsection{Datasets}

We used standard splits for training, development and validation data for the ConLL-2003 dataset \cite{conll-2003}. This is a named entity recognition dataset, which concentrates on four types of named entities: persons, locations, organizations and names of miscellaneous entities that do not belong to the previous three groups. The annotations are publicly available, but the corpus used in the English dataset are under the copyright of Reuters Ltd. The dataset covers two languages: English and German. Details of the number of examples and label distributions are shown in tables \ref{tab:english-conll2003} and \ref{tab:german-conll2003}. The dataset is available from \url{https://huggingface.co/datasets/conll2003}.

\begin{table}[htp]
\resizebox{\linewidth}{!}{%
\begin{tabular}{lllllll}
\hline
\textbf{English} & \textbf{Sentences} & \textbf{Tokens} & \textbf{LOC} & \textbf{MISC} & \textbf{ORG} & \textbf{PER} \\
\hline
Train    & 14,987 & 203,621 & 7140 & 3438 & 6321 & 6600 \\
Validation & 3,466  & 51,362  & 1837 & 922  & 1341 & 1842 \\
Test        & 3,684  & 46,435  & 1668 & 702  & 1661 & 1617 \\
\hline
\end{tabular}%
}
\caption{Number of examples and label distributions for English data in CoNLL-2003 dataset.}
\label{tab:english-conll2003}
\end{table}

\begin{table}[htp]
\resizebox{\linewidth}{!}{%
\begin{tabular}{lllllll}
\hline
\textbf{German} & \textbf{Sentences} & \textbf{Tokens} & \textbf{LOC} & \textbf{MISC} & \textbf{ORG} & \textbf{PER} \\
\hline
Train      & 12,705 & 206,931 & 4363 & 2288 & 2427 & 2773 \\
Validation & 3,068  & 51,444  & 1181 & 1010 & 1241 & 1401 \\
Test       & 3,160  & 51,943  & 1035 & 670  & 773  & 1195 \\
\hline
\end{tabular}%
}
\caption{Number of examples and label distributions for German data in CoNLL-2003 dataset.}
\label{tab:german-conll2003}
\end{table}

We extracted concept annotations for tokens from BERT Concept Net \cite{dalvi2020discovering}. The publicly available dataset consists of 635,000 English sentences and 1.7 million tokens which are annotated with 219 unique concept labels. We used annotations for the concept SEM:named\_entity:location and SEM:named\_entity:person using the given script, and take only the first 5000 lines of annotated sentences due to compute restrictions. The full dataset and scripts are available from 
\url{https://neurox.qcri.org/projects/bert-concept-net/bert-concept-net.html}

\subsection{Infrastructure and runtime}

Our experiments were run on an Nvidia Tesla A30 GPU card. On average, it took 1 hour to fine-tune the pretrained model for three epochs, identify and prune concept neurons, and retrain the model incrementally for a total of 8 epochs. It took 3.5 hours to rank neurons in the model by importance and identify the highest activation tokens for each neuron in the model.

\subsection{Alternatives to probeless method}

An alternative method for identifying top concept neurons in \S\ref{sec:probeless} is to use a linear probe. This involves training a logistic regression probe to predict concept annotations associated with a set of sentences from neuron activations \cite{dalvi2019neurox}. The top concept neurons is then be computed at increasing percentages of the weight mass and then accumulated in order to create a global ordering of neurons by saliency. 

We performed experiments with our method using both the probeless and linear probe method. From our results, both techniques yielded less than a 1\% variation in numerical values and did not influence the overall trend. However, the probeless method is preferred since it is less computationally expensive than the linear probe, and is free of probing limitations that could impact ranking quality \cite{antverg2021}.

\section{Analyzing concept recovery and redistribution after neuroplasticity}
\label{sec:appendix}

\subsection{Performance recovery}
\label{app:performance-recovery}

Tables \ref{tab:location-performance-scores} and \ref{tab:location-performance-scores-2} display how performance on overall named entity recognition changes throughout the remapping process for an example run on a DistilBERT model. Table \ref{tab:location-specific-scores} displays how performance on named entity recognition, specifically for the task of recognising location names, changes throughout the remapping process. There is a substantial decline in precision, recall and F1 score for location name recognition post-pruning, but performance recovers gradually throughout retraining.

\begin{table}[htp!]
\resizebox{\linewidth}{!}{%
\begin{tabular}{lccc}
\hline
\textbf{Stage} & \textbf{Precision} & \textbf{Recall} & \textbf{F1} \\
\hline
Base model & 0.931 & 0.940 & 0.936 \\
\textbf{Pruned} & \textbf{0.015} & \textbf{0.131} & \textbf{0.028} \\
Retrained (2 epochs) & 0.909 & 0.919 & 0.914 \\
Retrained (4 epochs) & 0.919 & 0.929 & 0.924 \\
Retrained (6 epochs) & 0.922 & 0.934 & 0.928 \\ 
Retrained (8 epochs) & 0.924 & 0.935 & 0.929 \\
\hline
\end{tabular}%
}
\caption{Precision, recall and F1 score on overall named entity recognition for DistilBERT throughout remapping, after pruning concept neurons related to location names. Note the dramatic drop in performance after pruning concept neurons, which confirms we have removed features that the model was using for predictions.}
\label{tab:location-performance-scores}
\end{table}

\begin{table}[htp!]
\resizebox{\linewidth}{!}{%
\begin{tabular}{lrrr}
\hline
\textbf{Stage} & \multicolumn{1}{l}{\textbf{Precision}} & \multicolumn{1}{l}{\textbf{Recall}} & \multicolumn{1}{l}{\textbf{F1}} \\
\hline
Base                 & 0.931 & 0.94  & 0.936 \\
\textbf{Pruned}      & \textbf{0.021} & \textbf{0.161} & \textbf{0.037} \\
Retrained (2 epochs) & 0.922 & 0.934 & 0.927 \\
Retrained (4 epochs) & 0.926 & 0.938 & 0.932 \\
Retrained (6 epochs) & 0.925 & 0.934 & 0.930  \\
Retrained (8 epochs) & 0.929 & 0.939 & 0.934 \\
\hline
\end{tabular}%
}
\caption{Precision, recall and F1 score on overall named entity recognition for DistilBERT throughout remapping, after pruning neurons related to location names in a separate run.}
\label{tab:location-performance-scores-2}
\end{table}

\begin{table}[htp!]
\resizebox{\linewidth}{!}{%
\begin{tabular}{lccc}
\hline
\textbf{Stage} & \textbf{Precision} & \textbf{Recall} & \textbf{F1} \\
\hline
Base model & 0.956 & 0.962 & 0.959 \\
\textbf{Pruned} & \textbf{0.039} & \textbf{0.067} & \textbf{0.049} \\
Retrained (2 epochs) & 0.935 & 0.947 & 0.941 \\
Retrained (4 epochs) & 0.948 & 0.954 & 0.951 \\
Retrained (6 epochs) & 0.952 & 0.960 & 0.956 \\ 
Retrained (8 epochs) & 0.949 & 0.960 & 0.955 \\
\hline
\end{tabular}%
}
\caption{Precision, recall and F1 score for DistilBERT specifically on recognising location names throughout remapping. Note the dramatic drop in performance after pruning concept neurons, and the subsequent recovery in performance throughout retraining.}
\label{tab:location-specific-scores}
\end{table}

\begin{table}[htp!]
\resizebox{\linewidth}{!}{%
\begin{tabular}{lccc}
\hline
\textbf{Stage} & \textbf{Precision} & \textbf{Recall} & \textbf{F1} \\
\hline
Base model & 0.392& 0.940 & 0.936 \\
\textbf{Pruned} & \textbf{0.486}& \textbf{0.131} & \textbf{0.028} \\
Retrained (2 epochs) & 0.434& 0.919 & 0.914 \\
Retrained (4 epochs) & 0.368& 0.929 & 0.924 \\
Retrained (6 epochs) & 0.423& 0.934 & 0.928 \\ 
Retrained (8 epochs) & 0.394& 0.935 & 0.929 \\
\hline
\end{tabular}%
}
\caption{Precision, recall and F1 score on overall named entity recognition for a DistilGPT2 model throughout remapping, after pruning concept neurons related to location names. Note the drop in performance after pruning concept neurons, which confirms we have removed features that the model was using for predictions.}
\label{tab:gpt2-performance-scores}
\end{table}

Likewise, table \ref{tab:gpt2-performance-scores} displays how performance on overall named entity recognition changes over one run for a DistilGPT2 model. Interestingly, performance for the model after pruning and retraining for 8 epochs is higher than the performance on the original fine-tuned model. This suggests neuroplasticity could potentially improve better performance.

\subsection{Complex concepts in later layers are redistributed to earlier ones}
\label{app:redistribution}

Table \ref{tab:hats-retraining} shows the highest activating tokens (HATs) for the most salient neurons in the model after retraining for two epochs. Upon comparing the HATs in the retrained model and the base model, we confirm that the pruned concept (location names) is remapped within the first two epochs. 

\begin{table}[!htp]
\resizebox{\linewidth}{!}{%
\begin{tabular}{lp{0.4\linewidth}p{0.5\linewidth}}
\hline
\textbf{Neuron} & \textbf{Base model HATs} & \textbf{Retrained model (2 epochs) HATs} \\
\hline
4606   & 106, Homer, 39, Says, 57                               & Aires, Civic, Lumpur, Clara, Hampshire               \\
4983   & Wilcox, Clintons, Navarro, Peterson, Wallis            & Straits, Town, Congo, Delhi, Hague                   \\
4867   & M20, Laws, Nafta, Flight, WTO                          & Ghana, Normandy, Thousand, Troy, Holiday             \\
4976   & Workers, BioDefense, Regenerative, Sodra, Regeneration & Buckinghamshire, Dove, Manitoba, Griffith, Middlesex \\
5134   & Everest, Odd, Grey, Trappist, Achievement              & Cricket, Northridge, Cornwall, Madero, Scotia \\
\hline
\end{tabular}%
}
\caption{Highest activating tokens for most salient neurons in model after pruning the concept of location names and retraining for two epochs, compared with their highest activating tokens in the base model.}
\label{tab:hats-retraining}
\end{table}

\begin{figure}[htp!]
\begin{subfigure}{\linewidth}
    \includegraphics[width=\linewidth,trim={0cm 0cm 0cm 1cm},clip]{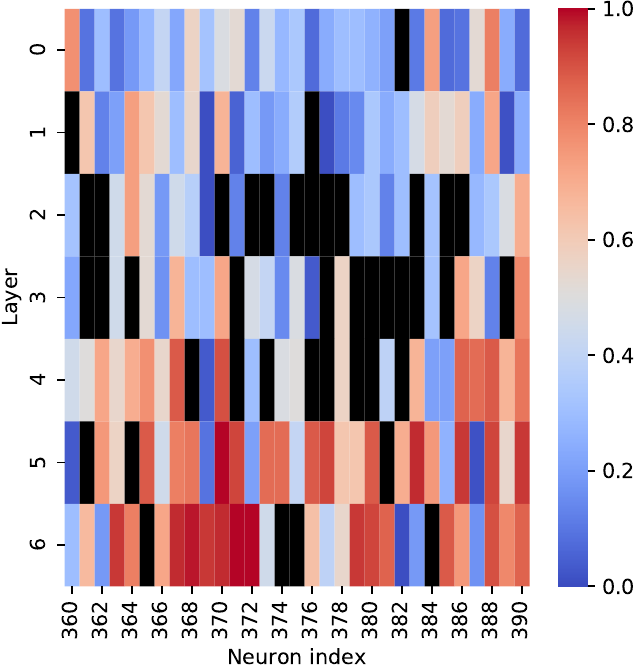}
    \caption{Concept saliency for neurons in model before pruning and retraining. Neurons in later layers are more salient.}
\end{subfigure}
\begin{subfigure}{\linewidth}
    \includegraphics[width=\linewidth,trim={0cm 0cm 0cm 1cm},clip]{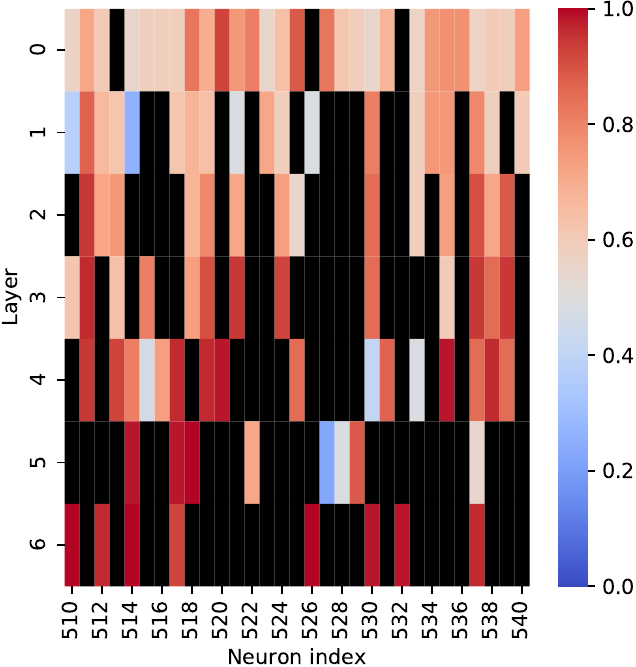}
    \caption{Concept saliency for neurons after retraining for 8 epochs. The most salient neurons are still in later layers, but earlier layers demonstrate increased saliency scores.}
\end{subfigure}
\caption{Representative sample of neurons in DistilBERT model before and after neuroplasticity occurs, highlighted by concept saliency for location names. Black units are neurons that were ablated.}
\label{fig:redistribution}
\end{figure}

Figure \ref{fig:redistribution} visualizes how concept saliency is redistributed from later layers to earlier layers in the middle of the DistilBERT model. Neurons located in later layers of the base model (layer 5 and 6) show higher concept saliency. Among all the concept neurons to be pruned from the base model, neurons in layers 5 and 6 have the highest saliency scores. This suggests that, in the base model, neurons in later layers vary more in the extent to which they capture the concept of location, but specific neurons represent the concept more strongly than the rest of the model.

\begin{table}[htp]
\centering
\resizebox{\linewidth}{!}{%
\begin{tabular}{llc}
\hline
\textbf{Layer} & \textbf{HATs of random neurons} & \textbf{Saliency} \\
\hline
1 & deepest, scary, Half, tests & 0.649 \\
2 & representation, hum, Lenfant, calm & 0.654 \\
3 & Bureau, \textbf{Durham}, Rookie, Christ & 0.564 \\
4 & \textbf{Quincy}, \textbf{Cove}, \textbf{Fulham}, Chorus & 0.747 \\
5 & Wilder, \textbf{Weisel}, Tour, \textbf{Salvador} & 0.792 \\
6 & \textbf{MA}, \textbf{Manchester}, Premiere, Placita & 0.829 \\
\hline
\end{tabular}%
}
\caption{Highest activation tokens (HATs) for random sample of neurons in different layers of the model, which was retrained after the concept of location was pruned. Tokens that could be associated with location are in bold. The pruned concept reappears not only in later layers, but also in middle layers 3 and 4. 
}
\label{tab:polysemantic-hats}
\end{table}

We verify that the pruned concept is redistributed by examining HATs for neurons across different layers after neuroplasticity. In Table \ref{tab:polysemantic-hats}, we randomly sample 4 neurons from each layer of the model - which was retrained after the concept of location names was pruned - and extract the top activating token for each neuron. As expected from the similar saliency scores, neurons in layers 3 and 4 now noticeably activate on location names in addition to neurons in layers 5 and 6. Interestingly, layers 3 and 4 also activate on tokens that could be seen as location words in one context but as a different concept in another (e.g. ``Quincy'' is both the name of a city and the name of a person). In contrast, neurons in later layers more consistently activate solely on location names. This pattern appears throughout the model.

This indicates that not only do middle layer neurons capture the pruned concept more strongly than before neuroplasticity occurred, but they also capture unrelated and/or context-dependent aspects of the pruned concept.

\subsection{Verifying our measure of concept saliency}
\label{app:verify-saliency}

To verify that our measure of concept saliency corresponds with the extent to which a concept is captured, we cross-check the HATs for neurons across different layers. We expect to see HATs in layers with higher concept saliency scores to be more associated with the concept of location. This is validated in Table 3, where we see that the HATs for neurons with the highest saliency are much more related to location names, compared to neurons in previous layers with lower saliency scores.

\begin{table}[htp!]
\resizebox{\linewidth}{!}{%
\begin{tabular}{lp{0.6\linewidth}l}
\hline
\textbf{Neuron} & \textbf{HATs}                                       & \textbf{Saliency} \\
\hline
1483            & aftermath, depiction, motto, constellation, kissing & 0.393             \\
2378            & mash, vandalized, Dortmund, denim, Kitty            & 0.526             \\
3754            & y, Relief, Charitable, Program, Haag                & 0.678             \\
1940            & rail, Chef, Sometimes, aspects, understanding       & 0.716             \\
4427            & Lithuanian, Australian, Gothic, Malaysian, Doctor   & 0.810             \\
4819            & Outback, Burger, Fords, Alignment, Universe         & 1.000     \\
\hline
\end{tabular}%
}
\caption{Highest activation tokens (HATs) for neurons with different concept saliency scores. Note that neurons with the highest concept saliency scores are clearly related to the concept of location, whereas neurons with low saliency are unrelated.}
\label{tab:verify-saliency}
\end{table}

\section{Generalizing concept remapping to person names}
\label{app:different-concept}
\subsection{Inducing concept remapping for the concept of person names}
\label{app:person-names}

To verify our overall results, we induce neuroplasticity for the concept of \textit{person names} in a DistilBERT model trained on NER. We find that a similar pattern emerges in how concepts are captured before and after remapping, and in where concepts reappear in terms of similarity.

\begin{table}[htp]
\resizebox{\linewidth}{!}{%
\begin{tabular}{lccc}
\hline
\textbf{Stage} & \textbf{Precision} & \textbf{Recall} & \textbf{F1} \\
\hline
Base model & 0.934 & 0.941 & 0.938 \\
\textbf{Pruned} & \textbf{0.021} & \textbf{0.150} & \textbf{0.037} \\
Retrained (2 epochs) & 0.897 & 0.913 & 0.905 \\
Retrained (4 epochs) & 0.910 & 0.923 & 0.917 \\
Retrained (6 epochs) & 0.911 & 0.928 & 0.919 \\ 
Retrained (8 epochs) & 0.913 & 0.928 & 0.921 \\
\hline
\end{tabular}%
}
\caption{Precision, recall and F1 score on overall named entity recognition throughout remapping, after removing concept neurons related to \textit{person names}. \label{table:concept-names}}
\label{tab:person-overall-scores}
\end{table}

\begin{table}[htp]
\resizebox{\linewidth}{!}{%
\begin{tabular}{lccc}
\hline
\textbf{Stage} & \textbf{Precision} & \textbf{Recall} & \textbf{F1} \\
\hline
Base model & 0.963 & 0.967 & 0.965 \\
\textbf{Pruned} & \textbf{0.011} & \textbf{0.091} & \textbf{0.020} \\
Retrained (2 epochs) & 0.941 & 0.954 & 0.947 \\
Retrained (4 epochs) & 0.941 & 0.957 & 0.949 \\
Retrained (6 epochs) & 0.941 & 0.959 & 0.950 \\ 
Retrained (8 epochs) & 0.937 & 0.961 & 0.949 \\
\hline
\end{tabular}%
}
\caption{Precision, recall and F1 score on recognising specifically person names throughout remapping, after removing concept neurons related to \textit{person names}. \label{table:person-concept-names}}
\label{tab:person-specific-scores}
\end{table}

After pruning top concept neurons related to person names, there is a significant decline in overall performance on named entity recognition, and performance on recognising person names, as shown in tables \ref{tab:person-overall-scores} and \ref{tab:person-specific-scores}. However, after retraining the model for 8 epochs, performance improves gradually such that the final retrained model's performance nearly matches its baseline. The dramatic drop and the subsequent recovery in performance provides further evidence for the feasibility of recovering performance despite the removal of crucial concept neurons.

\begin{figure}[!htb]
\begin{subfigure}{\linewidth}
    \includegraphics[width=\linewidth,trim={0cm 0cm 0cm 1cm},clip]{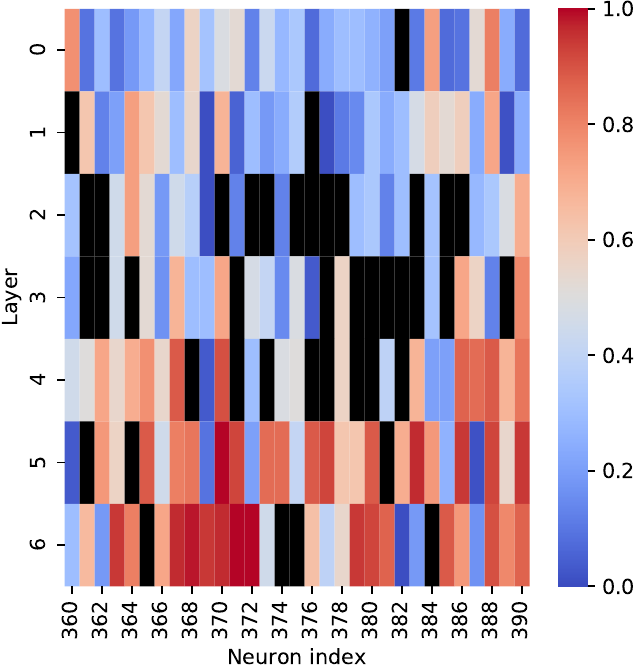}
    \caption{Concept saliency for neurons in model before pruning and retraining. Neurons in later layers are more salient.}
\end{subfigure}
\begin{subfigure}{\linewidth}
    \includegraphics[width=\linewidth,trim={0cm 0cm 0cm 1cm},clip]{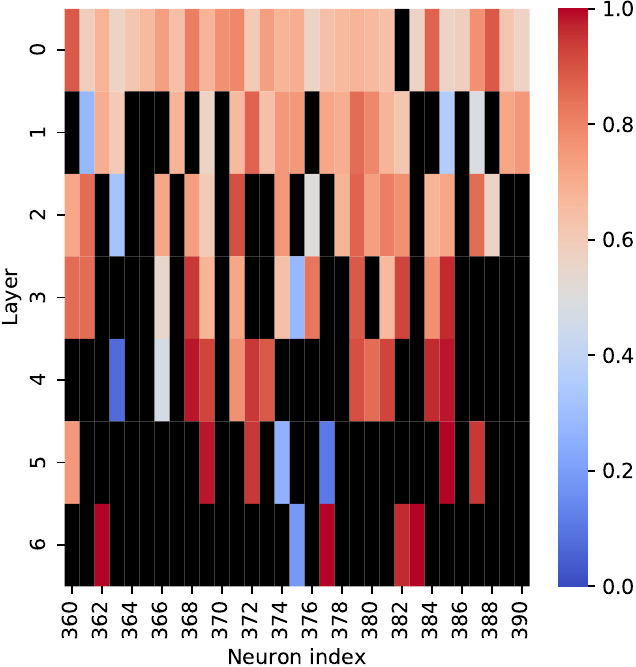}
    \caption{Concept saliency for neurons in model after retraining for 8 epochs. The most salient neurons are in layers 5 and 6, but earlier layers demonstrate higher saliency scores than before.}
\end{subfigure}
\caption{Representative sample of neurons in DistilBERT model before and after neuroplasticity occurs, highlighted by concept saliency for person names. A higher saliency means that the neuron captures the concept more strongly than other neurons. Black units are (pruned) neurons that do not activate on non-null tokens.}
\label{fig:redistribution-names}
\end{figure}

\subsection{Redistribution of concepts after pruning concept of person names}
\label{app:name-remapping-hats}

The distribution of concept saliency for person names is also similar to the distribution of saliency for location names. As seen from Figure \ref{fig:redistribution-names}, the concept of person names is captured most significantly in neurons in the later layers in the base model. The mean saliency scores in layer 5 and 6 in the base model (0.559 and 0.610, respectively) are considerably higher than the mean saliency scores in other layers. This reinforces the idea that higher level concepts such as location or people names is captured in later layers of the large language model. Interestingly, the concept of person names seems to be more concentrated in later layers of the model compared to the concept of location names. The range in mean concept saliency scores across layers for person names is 0.313, compared to 0.274 for location names.

After pruning and retraining, the concept of person names is also distributed more evenly over the middle and later layers of the model. This matches our results from \S\ref{section:redistribution}. As seen in figure \ref{fig:redistribution-names}, neurons in layers 5 and 6 remain the most salient (mean scores 0.797 and 0.872), but neurons in layers 3 and 4 also have noticeable saliency (mean scores 0.759 and 0.797) relative to other layers.

We confirm that neurons relearn both new concepts and old concepts over the retraining process, for the concept of person names. As the model is retrained, concept similarity does not change noticeably. Most neurons post-retraining capture new concepts which are very similar to concepts that they captured in the base model as shown in table \ref{tab:similarity-retraining-names}, particularly in the earlier layers. 

\begin{table}[htp]
\resizebox{\linewidth}{!}{%
\begin{tabular}{ccc}
\hline
\multicolumn{1}{c}{\multirow{2}{*}{\textbf{Layer}}} & \multicolumn{2}{c}{\textbf{Mean similarity over retraining}}                                                        \\
\multicolumn{1}{l}{}                                & \multicolumn{1}{c}{\textit{2 epochs}} & \multicolumn{1}{c}{\textit{8 epochs}} \\
\hline
0 & 0.925 & 0.925 \\
1 & 0.512 & 0.557 \\
2 & 0.430 & 0.444 \\
3 & 0.326 & 0.344 \\
4 & 0.186 & 0.209 \\
5 & 0.086 & 0.107 \\
6 & 0.040 & 0.043 \\
\hline
\end{tabular}%
}
\caption{Mean similarity scores across layers in the model at different stages after pruning the concept of person names (two epochs retraining versus 8 epochs retraining). Similarity is highest in middle layers. Similarity scores do not change significantly over retraining.}
\label{tab:similarity-retraining-names}
\end{table}

As a concrete example, table \ref{tab:hats-names} tracks the highest activating tokens for an example neuron throughout stages of remapping, after pruning neurons related to person names. A subset of HATs from the base model remain throughout remapping (``exist''), but the neuron learns to activate on new HATs related to the pruned concept.

\begin{table}[!htb]
\resizebox{\linewidth}{!}{%
\begin{tabular}{p{0.22\linewidth}p{0.6\linewidth}}
\hline
\textbf{Stage} & \textbf{HATs}\\
\hline
Base model & 
    \textbf{exist}, investment, halt, substitution, dog \\
Retrained (2 epochs) & 
    \textbf{exist}, Las, Hoover, discoveries, Spencer \\
Retrained (4 epochs) & 
    \textbf{exist}, Las, Hoover, Spencer, socially \\
Retrained (6 epochs) & 
    \textbf{exist}, Las, Hoover, Spencer, socially \\
Retrained (8 epochs) & 
    \textbf{exist}, Las, Hoover, Spencer, es \\

\hline
\end{tabular}%
}
\caption{Highest activating tokens (HATs) for neuron 2405 throughout stages of remapping. Note that the neuron eventually learns to activate on new tokens related to person names, as well as old tokens that it responded to in the base model.}
\label{tab:hats-names}
\end{table}

\subsection{Concept saliency versus similarity for the concept of person names}
\label{app:name-saliency-similarity}

\begin{figure}[!htb]
\includegraphics[width=\linewidth,trim={0.5cm 0cm 0.5cm 1.4cm},clip]{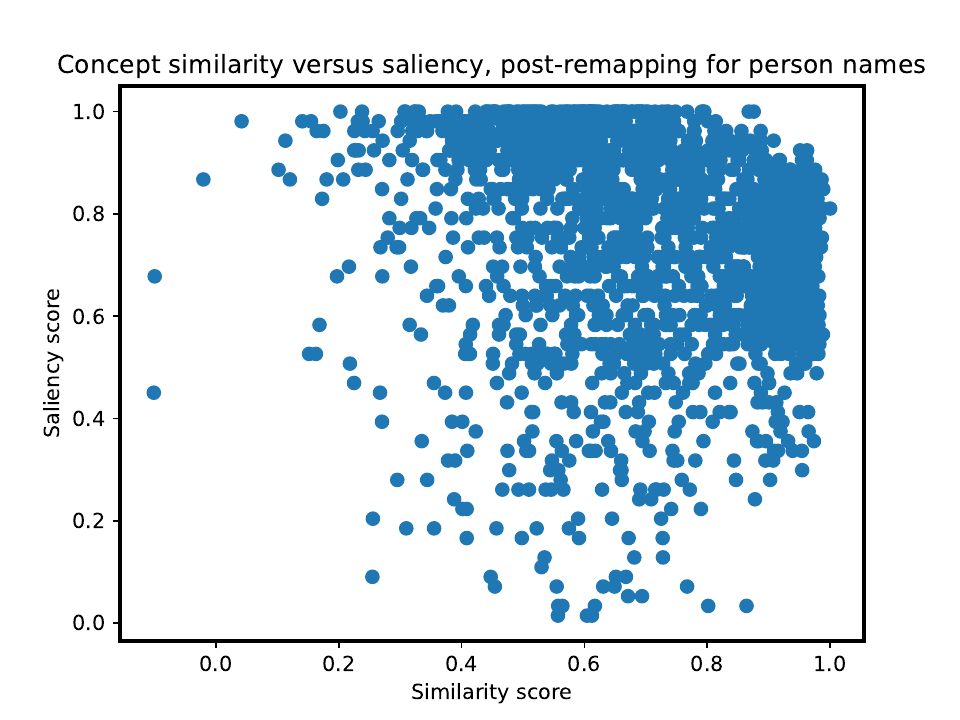}
\caption{In a model retrained for 8 epochs after the concept of person names was pruned, neurons with high saliency tend to also have high similarity. Neurons with negative similarity scores all had low saliency scores.}
\label{fig:similarity-saliency-person}
\end{figure}

Consider the same analysis as in \S\ref{section:similar-concepts} for the DistilBERT model after pruning the concept of person names and retraining as shown in figure \ref{fig:similarity-saliency-person}. The same pattern appears; there exists a dense cluster of neurons in the region of saliency above 0.5 and similarity above 0.5. Figure \ref{fig:similarity-saliency-person} reveals that most neurons for which the pruned concept of person names is remapped to originally activated strongly on a similar concept before neuroplasticity occurred. This provides further evidence to support a tentative correlation between concept similarity and saliency in DistilBERT.

\subsection{Pruning sub-concepts}
\label{app:prune-subconcepts}

To verify generalizability, we prune mutually exclusive subconcepts that are related to the same overall concept. We then check if the overall concept is relearned. In this example, we pruned the most salient 50\% of concept neurons related to the set of (mutually exclusive) concepts "SEM:origin:north\_america:usa", "SEM:origin:north\_america:mexico", and "SEM:origin:north\_america:canada" from a DistilBERT model. We then analyzed the model after retraining to detect if the overall concept "SEM:origin:north\_america" has been relearned.

\begin{figure}[htb!]
\includegraphics[width=\linewidth]{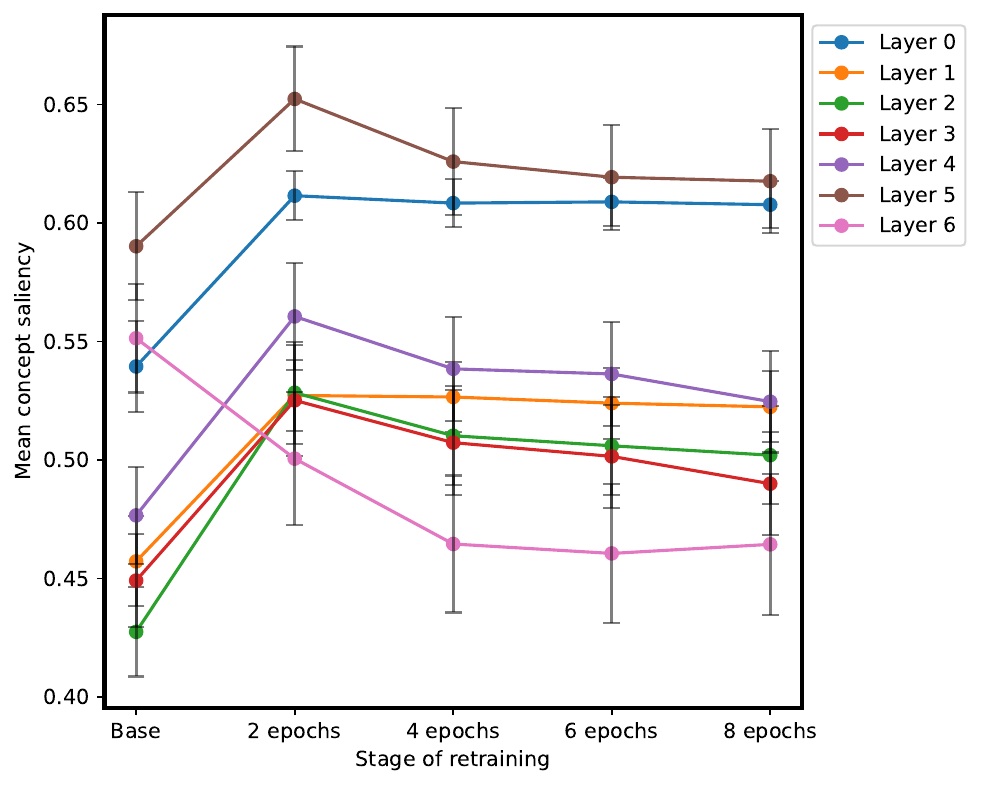}
\caption{Mean concept saliency for the concept of "North America", for neurons across different layers of DistilBERT throughout the process of neuroplasticity. Layer 6 shows a dramatic decrease in saliency, whereas layer 5 shows a large increase in concept saliency.}
\label{fig:saliency-layers-america}
\end{figure}

Figure \ref{fig:saliency-layers-america} shows that the concept is recovered after retraining, albeit not to a significant extent. Most layers have a final saliency score between 0.45 to 0.65, which is lower than scores in figures in \S\ref{section:redistribution}. However, layer 6 of the model shows a significant decrease in saliency score post-ablation, whereas layer 5 shows a signficant increase in saliency post-ablation. Representations of the overall concept for the pruned sub-concepts are therefore redistributed from later layers to earlier layers.

\section{Pseudocode algorithm}
\label{app:algo}
Algorithm \ref{alg:revised_algorithm} illustrates our method of inducing neuroplasticity and analyzing conceptual remapping.

\onecolumn
\begin{algorithm}
\caption{Neuroplasticity in LLMs}
\begin{algorithmic}[1]
\Require Pretrained Model $M$, Concept $C$
\Statex \textcolor{blue}{\# Begin with a pretrained model and a concept of interest}

\Statex
\Function{FineTune}{$M, C$}
    \State Fine-tune $M$ on task requiring $C$ \Comment{\textcolor{blue}{Adjust the model for a specific task}}
\EndFunction
\Statex

\Function{IdentifyConceptNeurons}{$M, C$}
    \State \( S \leftarrow \emptyset \) \Comment{\textcolor{blue}{Locate the neurons in the model}}
    \State Identify neurons $S$ for $C$
    \For{each neuron $s$ in $S$}
        \State Compute saliency of $s$ \Comment{\textcolor{blue}{Define how saliency is computed}}
    \EndFor
    \State \Return $S$
\EndFunction
\Statex

\Function{Prune}{$M, S$}
    \State Set weights in $S$ to zero \Comment{\textcolor{blue}{Reduce the influence of the concept}}
\EndFunction
\Statex

\Function{RetrainModel}{$M$}
    \State Retrain $M$ \Comment{\textcolor{blue}{Until recovery criteria met}}
    \State $R \leftarrow \emptyset$
    \State Identify neurons $R$
    \State \Return $R$
\EndFunction
\Statex

\Function{SaliencySimilarityAnalysis}{$M, S, R$}
    \For{each epoch}
        \If{epoch is even}
            \For{each neuron $r$ in $R$}
                \State Compute saliency of $r$ \Comment{\textcolor{blue}{Saliency metric for $r$}}
                \If{neuron $r$ is in $S$}
                    \State Compute similarity for $r$ \Comment{\textcolor{blue}{Compare $r$ with its original state in $S$}}
                \EndIf
            \EndFor
        \EndIf
    \EndFor
\EndFunction
\Statex
\end{algorithmic}
\label{alg:revised_algorithm}
\end{algorithm}

\section{Broader impact statement}
\label{app:sub-app-brd}
This study investigates neuroplasticity in large language models and its implications for model pruning, editing, transfer learning, and interpretability. The findings have the potential to improve the efficiency and safety of language models and increase public trust in large scale AI models. 

While this work aims to understand language models further to make them safer and aligned with user needs, better language model understanding and optimization may also lead to potential misuse or negative consequences. Enhanced model efficiency and capability, for instance, might be leveraged for malicious purposes, including spreading misinformation, creating more convincing deepfake content or even creation of dangerous pathogens. Given these potential wider impacts, we echo the need to approach AI research ethically and responsibly. Advancements in language model understanding, pruning, and editing should be accompanied by studying their ethical and safety consequences to create AI systems that are useful, safe, and aligned with human values.
\end{document}